\documentclass[10pt,twocolumn,letterpaper]{article}

\usepackage[pagenumbers]{cvpr} % To force page numbers, e.g. for an arXiv version

\usepackage[ruled]{algorithm2e}
\usepackage{bm}
\usepackage{multirow}
\usepackage{booktabs}
\usepackage{epsfig}
\usepackage[accsupp]{axessibility}
\usepackage{graphicx}
\usepackage{amsmath}
\usepackage{amssymb}
\usepackage{booktabs}
\usepackage[title]{appendix}
\usepackage{diagbox} 
\usepackage[pagebackref,breaklinks,colorlinks]{hyperref}

\usepackage[capitalize]{cleveref}
\crefname{section}{Sec.}{Secs.}
\Crefname{section}{Section}{Sections}
\Crefname{table}{Table}{Tables}
\crefname{table}{Tab.}{Tabs.}

\newcommand{\ytyn}[1]{\textcolor{black}{#1}}

\newcommand{\CUT}[1]{}

\def\cvprPaperID{8692} % *** Enter the CVPR Paper ID here
\def\confName{CVPR}
\def\confYear{2022}

\begin{document}

%%%%%%%%% TITLE - PLEASE UPDATE
\title{SWEM: Towards Real-Time Video Object Segmentation with Sequential Weighted Expectation-Maximization}

\author{Zhihui Lin${^{1*}}$, Tianyu Yang${^{2}}$, Maomao Li${^2}$, Ziyu Wang${^3}$, Chun Yuan${^{4\dag}}$, Wenhao Jiang${^3}$, and Wei Liu${^3}$ \\
${^1}$Department of Computer Science and Technologies, Tsinghua University, Beijing, China \\
${^2}$Tencent AI Lab, Shenzhen, China\quad ${^3}$Tencent Data Platform, Shenzhen, China \\
${^4}$Tsinghua Shenzhen International Graduate School, Peng Cheng Lab, Shenzhen, China \\
{\tt\small \{lin-zh14@mails,\ yuanc@sz\}.tsinghua.edu.cn} \quad {\tt\small tianyu-yang@outlook.com} \\
{\tt\small \{limaomao07,\ cswhjiang\}@gmail.com \quad wangziyukobe@163.com \quad wl2223@columbia.edu}
}

\maketitle

%%%%%%%%% ABSTRACT
\begin{abstract}
   Matching-based methods, especially those based on space-time memory, are significantly ahead of other solutions in semi-supervised video object segmentation (VOS). However, continuously growing and redundant template features lead to an inefficient inference. To alleviate this, we propose a novel \textbf{S}equential \textbf{W}eighted \textbf{E}xpectation-\textbf{M}aximization (SWEM) network to greatly reduce the redundancy of memory features. Different from the previous methods which only detect feature redundancy between frames, SWEM merges both intra-frame and inter-frame similar features by leveraging the sequential weighted EM algorithm. Further, adaptive weights for frame features endow SWEM with the flexibility to represent hard samples, improving the discrimination of templates. 
   Besides, the proposed method maintains a fixed number of template features in memory, which ensures the stable inference complexity of the VOS system. Extensive experiments on commonly used DAVIS and YouTube-VOS datasets verify the high efficiency (36 FPS) and high performance (84.3\% $\mathcal{J}\&\mathcal{F}$ on DAVIS 2017 validation dataset) of SWEM. Code is available at: \url{https://github.com/lmm077/SWEM}.
\end{abstract}

\let\thefootnote\relax\footnotetext{*Work done during an internship at Tencent AI Lab}
\let\thefootnote\relax\footnotetext{$\dag$Corresponding Author} 

%%%%%%%%% BODY TEXT
\section{Introduction}
Semi-supervised video object segmentation (VOS) has seized great interest recent years\cite{caelles2017one,perazzi2017learning,Johnander_2019_CVPR,Voigtlaender_2019_CVPR,Li_2019_ICCV,Oh_2019_ICCV,Wang_2019_ICCV,seong_2020_ECCV,lu_2020_ECCV,liang2020video,hu2021learning,xie2021efficient,wang2021swiftnet,duke2021sstvos,yang2021associating, seong2021hierarchical,cheng2021stcn} in the computer vision community.
It aims to segment the objects of interest from the background in a video, where only the mask annotation of the first frame is provided during testing.  A group of early methods concentrate on on-line fine-tuning\cite{caelles2017one,bao2018cnn,maninis2018video,luiten2018premvos,khoreva2017lucid} with the first annotated frame. However, these method tends to suffer from model degradation caused by target appearance changes as video goes on. Besides, propagation-based methods use masks computed in previous frames to estimate masks in the current frame\cite{perazzi2017learning,cheng2018fast,wug2018fast,wang2018semi}, which is, however, vulnerable to occlusions and rapid motion.

\begin{figure}
\centering
  \begin{subfigure}{0.48\linewidth}
  \centering
      \includegraphics[width=0.95\textwidth]{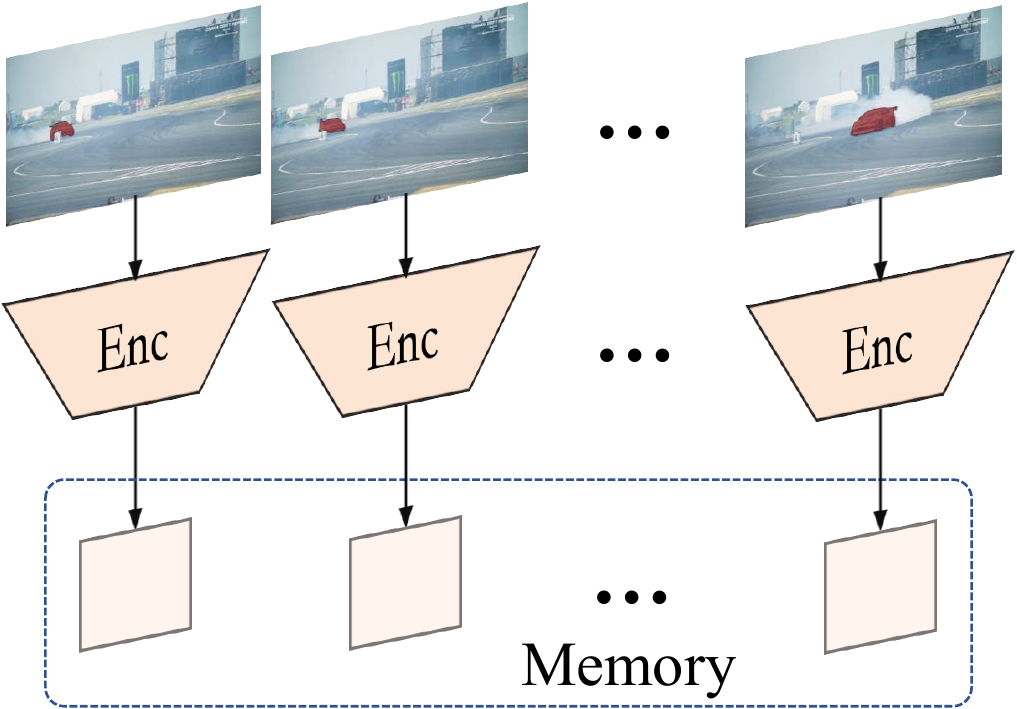}
    \caption{STM}\label{subfig:stm_mem}
  \end{subfigure}
  \hfill
  \begin{subfigure}{0.48\linewidth}
  \centering
      \includegraphics[width=0.95\textwidth]{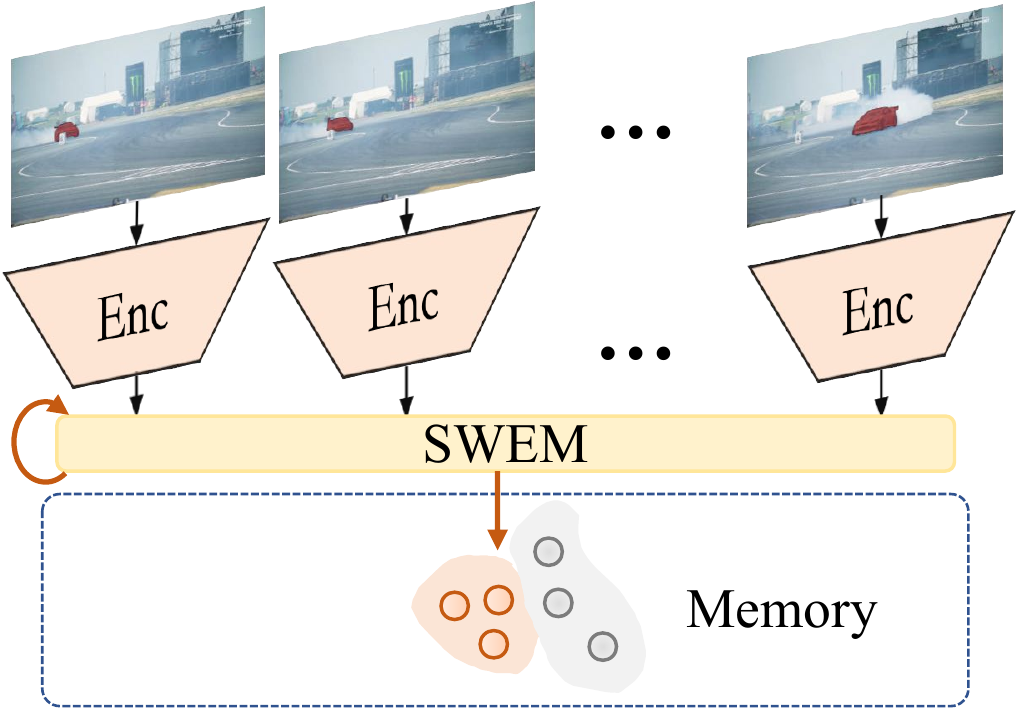}
    \caption{SWEM}\label{fig:subfig:swem_mem}
  \end{subfigure}
%\subfigure[STM]{
%\label{subfig:stm_mem} %% 第一幅图的标签
%\includegraphics[width=0.20\textwidth]{figures/stm_mem.pdf}}
%\subfigure[SWEM]{
%\label{fig:subfig:swem_mem} %% 第二幅图的标签
%\quad
%\includegraphics[width=0.20\textwidth]{figures/swem_mem.pdf}}
\vspace{-0.3 cm}
\caption{Instead of storing all past frames features as memory, just like STM~\cite{Oh_2019_ICCV} and following methods~\cite{seong_2020_ECCV,hu2021learning,seong2021hierarchical,cheng2021stcn} do, our SWEM sequentially updates a compact set of bases with a fixed size, greatly reducing the inter-frame and intra-frame redundancy.}
\vspace{-0.5 cm}
\label{fig:stm_swem_mem} %% label for entire figure
\end{figure}

Recently, matching-based VOS methods~\cite{chen2018blazingly,hu2018videomatch,Voigtlaender_2019_CVPR,lu_2020_ECCV,yang_2020_ECCV,seong_2020_ECCV,liang2020video, hu2021learning,xie2021efficient,wang2021swiftnet,seong2021hierarchical,cheng2021stcn} have achieved striking performance. Such matching-based methods first exploit previous frames to construct target templates, and then calculate the pixel-level correlations between the new coming frame embeddings and the target templates to perform the segmentation.
As seen in Figure~\ref{fig:stm_swem_mem}, the Space-Time Memory Network (STM)~\cite{Oh_2019_ICCV} and the following STM-like methods ~\cite{seong_2020_ECCV,hu2021learning,xie2021efficient,seong2021hierarchical,cheng2021stcn} leverage memory networks to store template features every T frames endlessly, which is prone to missing key-frame information and running out of memory for long-term videos. Besides, given that the inter-frame redundancy of video features would harm the efficiency of matching, another group of methods AFB\_URR~\cite{liang2020video} and Swift~\cite{wang2021swiftnet} take advantage of the similarity of inter-frame features to selectively update partial features. Nonetheless, they all fail to balance the performance and efficiency through a hand-crafted similarity threshold.

%尽管之前的工作已经取得了很好的结果，我们发现不仅是帧间冗余，帧内冗余也是导致模版匹配效率不足的一个重要因素。
Although past efforts have achieved promising results, we argue that both inter-frame redundancy and intra-frame one pose the main obstacles that prevent efficient template matching.
Here comes to a question that can we achieve a real-time VOS system by considering both the inter-frame and intra-frame redundancy simultaneously? In this paper, we will explore its feasibility.

Inspired by the Expectation-Maximization Attention (EMA) ~\cite{Li_2019_ICCV}, we intend to construct a set of low-rank bases for memory features through Expectation-Maximization (EM)~\cite{em} iterations. Here, the number of bases is far less than that of image pixels. Thus, bases can be regarded as a more compact representation, which can greatly reduce the intra-frame redundancy. Instead of applying EM directly, we adopt Weighted Expectation-Maximization (WEM) with the predicted mask as the \textbf{fixed} weights to explicitly construct foreground and background bases in each frame. What's more, we also propose a weighted EM with \textbf{adaptive} weights, which give larger weights for hard samples during generating bases. Here, the hard samples refer to those pixels that are not well expressed by bases, but are important for object segmentation.

WEM can deal with the intra-frame redundancy effectively; however, inter-frame one remains unsolved. Applying WEM on a single frame is efficient, but the computation complexity will be dramatically increased if directly applying it to all growing memory features. To further reduce the inter-frame redundancy, we propose the Sequential Weighted Expectation-Maximization (SWEM), where features of only one frame participate in the EM iterations during the memory updating stage. The memory bases will be updated with the new frame features through similarities rather than a simple linear combination. Formally, this updating process is equivalent to a weighted average of all past frame features. 
As shown in Figure~\ref{fig:stm_swem_mem}, 
compared with STM~\cite{Oh_2019_ICCV} which saves all historical frame information as the memory template of objects, our SWEM only updates a more compact set of bases sequentially, thus greatly reducing the inter-frame and intra-frame redundancy.

Our contributions can be summarized as follows:
\begin{itemize}
\item 
\vspace{-0.2cm}
We propose a fast and robust matching-based method for VOS, dubbed Sequential Weighted Expectation-Maximization (SWEM) network,
where a set of compact bases are constructed and updated sequentially, reducing both the inter- and intra-frame redundancy.
\item
\vspace{-0.2cm}
We introduce an adaptive weights calculation approach for weighted EM, which makes the base features pay more attention to hard samples.
\item
\vspace{-0.2cm}
 Without bells and whistles, SWEM reaches a level close to state-of-the-art performance, while maintaining an inference speed of 36 FPS.
\end{itemize}

%%%%%%%%%%%%%%%%%%%%%%%%%%%%%%%%%%%%%%%%%%%%%%%%%%%%%%%%%%%%%%%%%%%%%%%%%%%%%%%%%%%%%%%%%%%%%%%%%%%%%%%%%%%%%%%%%%%%%%% Related Work

\section{Related Work}
\noindent{\textbf{Matching-based Methods for VOS.}} 
Recent years have seen a surge of interest in video object segmentation under the semi-supervised setting.  A number of matching-based methods~\cite{wang2019fast,hu2018videomatch,chen2018blazingly,Wang_2019_ICCV,Oh_2019_ICCV,seong_2020_ECCV,lu_2020_ECCV,liang2020video, hu2021learning,xie2021efficient,wang2021swiftnet,seong2021hierarchical,cheng2021stcn} regard the first or intermediate frames as a target template, which is then used to match the pixel-level feature embedding in the new frame. To obtain both long-term and short-term object appearance information, FEELVOS~\cite{Voigtlaender_2019_CVPR} and CFBI~\cite{yang_2020_ECCV} match the current frame with both the first frame and the previous frame to obtain both global and local temporal dependencies. Besides, STM~\cite{Oh_2019_ICCV} and the following methods~\cite{seong_2020_ECCV, hu2021learning,xie2021efficient,seong2021hierarchical,cheng2021stcn} store multiple memory templates from all previous frames as templates, which is redundant and time-consuming during matching. In contrast, we propose a novel method named SWEM, which only stores a set of low-rank and updated basis features for each target, making the target representation more compact and efficient. 
 
\noindent{\textbf{Learning Fast and Robust VOS.}} 
Leaning a fast and robust model is a common goal since both accuracy and speed are important in practical applications~\cite{Wang_2019_ICCV,wang2019fast,Robinson_2020_CVPR,Chen_2020_CVPR,Zhang_2020_CVPR,li_2020_ECCV_GCM}. RANet~\cite{Wang_2019_ICCV} only uses the first frame as the target template for an acceptable speed. As tracker-based methods, SiamMask~\cite{wang2019fast} and SAT~\cite{Chen_2020_CVPR} only process the region of interest. TVOS~\cite{Zhang_2020_CVPR} directly propagates target masks according to feature similarity in the embedding space. In general, to achieve fast VOS, the previous methods sacrificed the integrity of the target representation, which substantially degrades segmentation performance. Swift~\cite{wang2021swiftnet} uses a variation-aware trigger module to compute the inter-frame difference to update frames with diverse dynamics. Further, only partial features that are significantly different from memory features will be updated. 

In this work, we consider reducing inter- and intra-frame redundancy simultaneously. The proposed weighted EM greatly reduces the intra-frame redundancy by iteratively constructing compact base features for the whole frame. To diminish the inter-frame redundancy, we further extend weighted EM in a sequential manner,
which can adaptively update the model without increasing the number of matching templates, thus making our model fast and robust.

\section{Preliminaries}

\subsection{Expectation-Maximization Algorithm}
The expectation-maximization (EM)~\cite{em} is an iteration-based algorithm, which can be used to estimate parameters of latent variable modes by maximizing the likelihood. The task is defined as estimating model parameters $\theta$ based on observation data set $\mathbf{X}$ %=\{\mathbf{x}_i\}_{i=1}^{N}$ 
and corresponding latent variables $\mathbf{Z}$. %=\{\mathbf{\ytyn{z}}_i\}_{i=1}^{N}$. 
Each EM iteration involves two steps, the Expectation step (E step) and the Maximization step (M step). At the $r$-th iteration, E step finds the posterior $P(\mathbf{Z}|\mathbf{X}, \theta^{r-1})$ and computes the expectation:
\vspace{-0.2cm}
	\begin{equation}
	\begin{split}
	 \mathcal{Q}(\theta, \theta^{r-1})=\sum P(\mathbf{Z}|\mathbf{X}, \theta^{r-1})\ln P(\mathbf{X}, \mathbf{Z}|\theta).
	\end{split}
	\label{eq:EM_E}
	\end{equation}
M step estimates parameters by maximizing the above data likelihood:
\vspace{-0.2cm}
	\begin{equation}
	\begin{split}
	 \theta^{r} = {\arg\max}_{\theta}\mathcal{Q}(\theta, \theta^{r-1}).
	\end{split}
	\label{eq:EM_M}
	\end{equation}
The E step and the M step are alternately executed $R$ times to achieve the convergence criterion.

\subsection{Expectation-Maximization Attention}
Expectation-Maximization Attention (EMA)~\cite{Li_2019_ICCV} is proposed to formulate the attention mechanism~\cite{nonlocal} into an expectation-maximization manner. Specifically, instead of regarding \emph{all pixels} as reconstruction bases, EMA iteratively estimates a much more compact set of bases for each image. EMA consists of three steps, namely \textit{Responsibility Estimation} (RE), \textit{Likelihood Maximization} (LM), and \textit{Data Re-estimation} (DR). Denote $\ytyn{\mathbf{X}=\{\mathbf{x}_n\}_{n=1}^{N}}\in \mathbb{R}^{N\times C}$ as image feature, $\ytyn{\mathcal{M}=\{\bm{\mu}_k\}_{k=1}^K}\in \mathbb{R}^{K\times C}$ as the randomly initialized base features, where $N$, $C$, and $K$ indicate the numbers of pixels, channels, and bases. RE estimates the hidden variable $\ytyn{\mathbf{Z} = \{z_{nk}\}_{n=1,k=1}^{N,K}}\in \mathbb{R}^{N\times K}$, where the responsibility $z_{nk}$ represents the probability of the $n$-th pixel belonging to the $k$-th base:
\vspace{-0.2cm}
	\begin{equation}
	\begin{split}
	 \ytyn{z_{nk}}=\frac{\exp(\mathbf{x}_{n}\bm{\mu}^{\top}_{k}/\tau)}{\sum_{j=1}^{K}\exp(\mathbf{x}_{n}\bm{\mu}^{\top}_{j}/\tau)}.
	\end{split}
	\label{eq:EMA_AE}
	\end{equation}
Here, $\tau$ is a hyper-parameter which controls the shape of distribution $\mathbf{Z}$. Then, LM updates base features $\ytyn{\mathcal{M}}$ by applying the weighted average on feature $\mathbf{X}$. That is, the $k$-th base is updated by:
\vspace{-0.2cm}
\begin{equation}
 \bm{\mu}_{k} = \frac{\sum^{N}_{n=1}\ytyn{z_{nk}}\mathbf{x}_{n}} {\sum^{N}_{n=1}\ytyn{z_{nk}}}.
 \label{eq:EMA_AM}
\end{equation}
Note that RE and LM are iteratively executed $R$ times until convergence. Finally, {DR} reconstructs a low-rank version of $\mathbf{X}$ using $\tilde{\mathbf{X}}=
\mathbf{Z}\ytyn{\mathcal{M}}$. Since $K$ is much less than $N$, basis set $\ytyn{\mathcal{M}}$ can be treated as a compact representation for image feature $\mathbf{X}$. Inspired by EMA, {we} consider replacing redundant memory features with more compact base features.

\begin{figure}
\begin{center}
\includegraphics[width=1.0\linewidth]{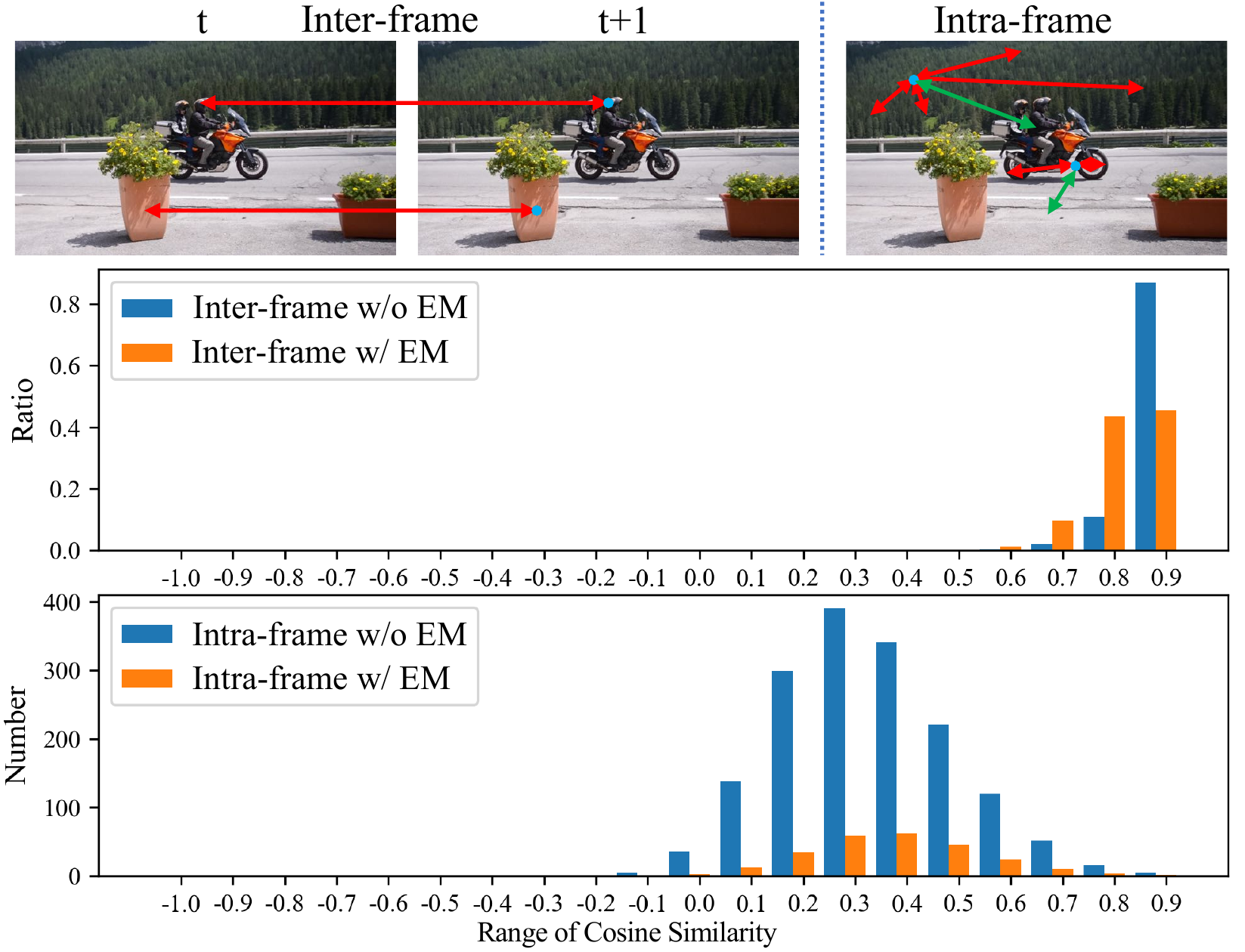}
\end{center}
\vspace{-0.5cm}
   \caption{The illustration of inter-frame and intra-frame redundancy of video features.}
\label{fig:inter_intra}
\vspace{-0.5cm}
\end{figure}

\subsection{Redundancy of the Space-time Memory}
To get a more intuitive understanding of feature redundancy in videos, we evaluate the inter-frame and intra-frame cosine similarities of videos features on the DAVIS 2017~\cite{pont20172017} validation set using the image encoder of STM~\cite{Oh_2019_ICCV} as the feature extractor. For each pixel in the current frame, we first calculate its maximum similarity with all pixels in the previous frame. In this way, $N$ such maximum similarities can be obtained. In Figure~\ref{fig:inter_intra}, we list the histogram of the maximum similarities, where the horizontal coordinate is the similarity range. Most of the similarities are larger than $0.6$, and nearly 87\% of similarities are larger than $0.9$, indicating high inter-frame redundancy in video sequences. In contrast, computing the maximum similarity for intra-frame redundancy measurement is not appropriate since spatial continuity would make most maximum similarities exceed $0.9$. Thus, we calculate all pair-wise similarities and count the average pair number of each frame under different similarities. The third line in Figure~\ref{fig:inter_intra} shows the statistics.  Most of the similarities between the two pixels in an image are positive, and more than 70\% of them are larger than $0.3$, which demonstrates the ubiquity of intra-frame redundancy.

To verify that the EM algorithm can find a more compact representation for image features and thus restrain the frame redundancy, we calculate the inter-frame and intra-frame similarity with a basis set rather than the entire image feature, where the basis set consists of 256 bases evaluated via EM iterations. Specifically, instead of calculating similarities between inter-frame bases, we calculate the maximum similarity between each frame feature and the base feature of the previous frame. As seen in Figure~\ref{fig:inter_intra}, more than 99\% of inter-frame similarities are larger than $0.7$. That is, although each frame has only 256 base features, which is far less than pixel number, it still meets the need of inter-frame matching. As for intra-frame similarities, although the similarity distribution is basically the same as that of the whole image feature, the number of large similarities has decreased significantly, which demonstrates the intra-frame redundancy is greatly reduced via EM iterations.

%%%%%%%%%%%%%%%%%%%%%%%%%%%%%%%%%%%%%%%%%%%%%%%%%%%%%%%%%%%%%%%%%%%% Figure inter-intra redundancy

\begin{figure*}
\begin{center}
\includegraphics[width=0.7\linewidth]{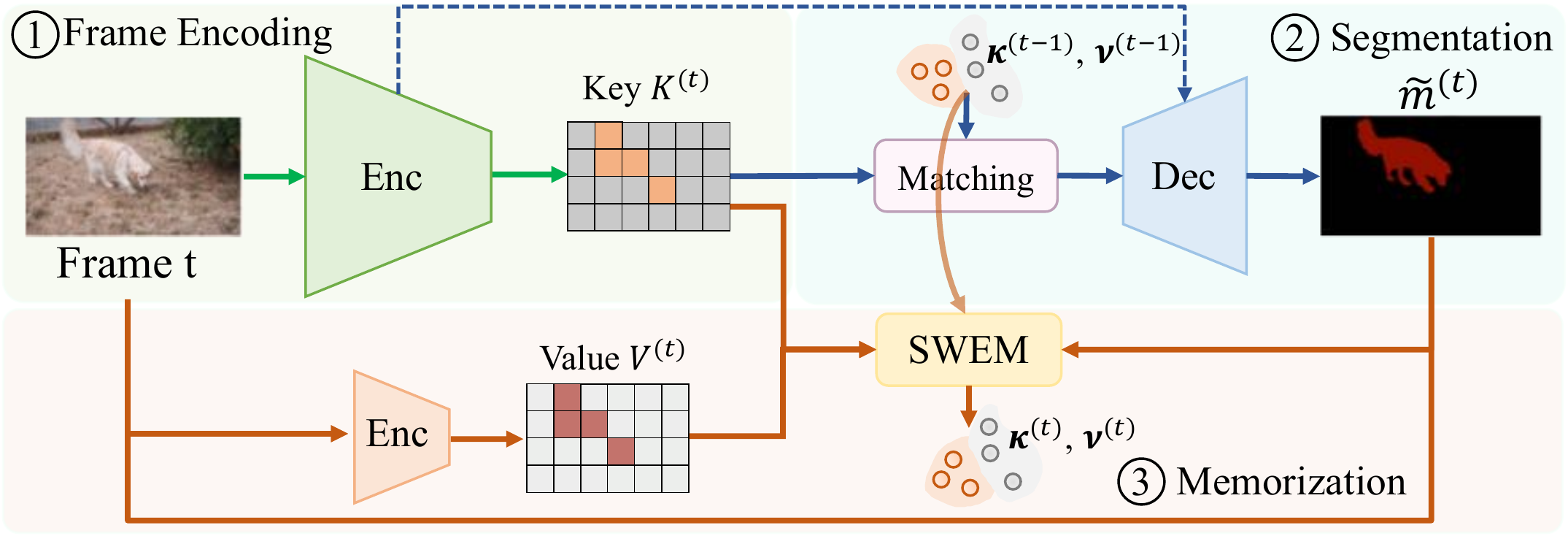}
\end{center}
\vspace{-0.5cm}
   \caption{
The matching-based pipeline of SWEM. The backbone network receives the $t$-th frame to capture general image features as \textbf{Key} $\mathbf{K}^{(t)}$. The features are used to match with target-specific memories. Through the matching process, the re-aggregated value and similarity map are obtained to be the target features for the final segmentation. Multi-level skip connections help refine the segmentation results for low to high resolution. The predicted mask is then employed with the intermediate image features to update bases via our SWEM.}
\label{fig:SWEM}
\vspace{-0.5cm}
\end{figure*}

\section{Proposed Approach}
\label{sec:apporach}
We first introduce the weighted EM, which leverages the predicted mask as weights to explicitly construct foreground and background bases separately in each frame. 
Furthermore, the \textbf{adaptive} weights make the model pay more attention to hard samples to improve the segmentation performance. Then, the core part of this work, the SWEM algorithm is detailed, which shows how to convert \textbf{growing} frame features into \textbf{fixed-size} bases. Finally, we describe the matching-based pipeline of the proposed SWEM. 

\subsection{Weighted Expectation-Maximization}
\label{subsec:WEM}
Although we have proved that using EM to find a more compact representation can reduce both inter and intra-frame redundancy, we argue that naively employing EM to learn a set of bases for memory features is not a reasonable solution in the VOS system. The reason here is that the memory bases would mix with both the foreground and background, which is unfavorable for object segmentation.
Instead, it is desirable to build low-rank foreground and background bases separately. To this end, we leverage Weighted Expectation-Maximization (WEM)~\cite{long2006spectral,tseng2007penalized,ackerman2012weighted,feldman2012data,gebru2016algorithms}, which is widely used for weighted data clustering. When using WEM to produce bases for images, Eq.~\eqref{eq:EMA_AM} would be modified as:
\begin{equation}
 \bm{\mu}_{k} = \frac{\sum^{N}_{n=1}\ytyn{z_{nk}}w_n\mathbf{x}_{n}} {\sum^{N}_{n=1}\ytyn{z_{nk}}w_n},
 \label{eq:WEM_M}
\end{equation}
where $w_n$ is the weight for $\mathbf{x}_{n}$. It is equivalent to ``seeing the $n$-th feature $w_n$ times"~\cite{gebru2016algorithms}.
Note that WEM makes it possible to construct separate foreground and background bases for template matching, where
the object foreground mask and the background mask of each frame can be used as the corresponding \textit{fixed} weights to substitute $w_n$. 
In this way, any irregular target region can be represented by a set of bases with a fixed size, which greatly reduces intra-frame redundancy.

%由于基特征的数量远少于帧特征，于是在匹配过程中，键匹配丢失的问题会更加严重，从而引起分割目标缺失的问题。为了缓解这一问题，我们在构建基的过程中通过自适应调整参与特征的权重来保障他们得到足够的匹配值。这里我们假设网络解码器的输出足够正确，并以此为标准，检查每个点的与基的匹配是否正确。

The essence of using WEM for compact representation learning is to perform clustering on all pixels in the image, and make different bases to represent each pixel. Because the number of bases is far less than that of pixels, the constructed segmentation target template would be incomplete to some extent and even be faced with the target lost situation. The expression degree of each pixel is different during basis set construction. Some pixels have little contribution for bases, but are very important for object segmentation, which are so-called hard samples.
To ensure that the hard samples could be assigned larger weights during basis set construction, we propose to \textit{adaptively} adjust the weights of pixels rather than directly employing the \textit{fixed} weights calculated via foreground and background masks.

We estimate a confidence score for each pixel by a foreground-background binary classification. Specifically, after the E-step of WEM iteration, each pixel is classified by the foreground or background bases, the classification probability of the $n$-th pixel can be calculated as:
\begin{equation}
\begin{split}
P^{fg}(\mathbf{x}_n) &= \frac{\sum_{k=1}^{K} \mathcal{K}(\mathbf{x}_n, \bm{\mu}^{fg}_k)}{\sum_{k=1}^{K} [\mathcal{K}(\mathbf{x}_n, \bm{\mu}^{fg}_k)+ \mathcal{K}(\mathbf{x}_n, \bm{\mu}^{bg}_k)]}, \\
P^{bg}(\mathbf{x}_n) &= 1 - P^{fg}(\mathbf{x}_n),
\end{split}
\label{eq:adapt_p}
\end{equation}
where $\ytyn{\bm{\mu}^{fg}_k}$ and $\ytyn{\bm{\mu}^{bg}_k}$ are foreground and background bases separately. $\mathcal{K}(\cdot)$ is a kernel function for calculating the similarity of two input features. Specifically, $\mathcal{K}(\mathbf{a}, \mathbf{b})=\exp(\frac{\mathbf{a}\mathbf{b}^{\top}/\tau}{\|\mathbf{a}\|\cdot \|\mathbf{b}\|})$. Eq.~(\ref{eq:adapt_p}) can be treated as a coarse segmentation result, since it provides the result of whether each pixel corresponds to the foreground or the background. Besides, the final segmentation (i.e., the output of the network decoder) can be obtained, which is considered more accurate than the coarse one since it is additionally constrained by ground-truth annotations. If coarse segmentation of a pixel is consistent with the final one, this pixel can be regarded as an easy sample. Otherwise, it would be regarded as a hard sample. We believe that the inconsistency of hard samples is because these pixels are neglected during the base construction process, which makes them struggle to achieve the same result as the final segmentation.
Supposing that $m^{fg}$ and $m^{bg}$ are soft masks of final segmentation, the \textit{adaptive weights} are estimated by:
\begin{equation}
\begin{split}
w^{fg}_n &= m^{fg}_n P^{bg}(\mathbf{x}_n),\quad w^{bg}_n = m^{bg}_n P^{fg}(\mathbf{x}_n).
\end{split}
\label{eq:adapt_w}
\end{equation}
The more inconsistent the coarse and final segmentation are, the higher weights would be given for base construction.

\subsection{Sequential Weighted EM}
To reduce inter-frame redundancy, previous methods  ~\cite{liang2020video,wang2021swiftnet} set a similarity threshold to ignore or merge similar features between frames. However, the segmentation performance and computational complexity are sensitive to this hand-crafted threshold. What's worse, it is hard to find an appropriate threshold to make a trade-off between performance and complexity. In this paper, we extend WEM in a sequential manner, yielding a Sequential Weighted EM (SWEM) algorithm to reduce both intra- and inter-frame redundancy without any threshold hyperparameter.

At the time step $t$, the ideal solution is to apply WEM to all previous $t-1$ frames for basis set construction.  
However, the growing scale of computing is unacceptable, which is not feasible for long-term segmentation. Therefore, we introduce a sequential weighted average of frame features when computing base assignment,  where the weights are estimated with time-dependent responsibility $\mathbf{Z}^{(t)}$.   
%%%%%%%%%%%%%%%%%%%%%%%%%%%%%%%%%%%%%%%%%%%%%%%%%%%%%%%%%%%%%%%%%%%%%%%%%%%%%%%%%%%%% Algorithm SWEM

%%%%%%%%%%%%%%%%%%%%%%%%%%%%%%%%%%%%%%%%%%%%%%%%%%%%%%%%%%%%%%%%%%%%%%%%%%%%%%%%%%%%% Algorithm 
Concretely, we extend the WEM sequentially and reformulate Eq. (\ref{eq:WEM_M}) as:
%\vspace{-0.2cm}
\begin{equation}
    \bm{\mu}^{(t)}_k = \frac{\sum_{i=1}^{t}\sum_{n=1}^{N}\ytyn{{z}^{(t)}_{nk}}w^{(t)}_n\mathbf{x}^{(t)}_n}{\sum_{i=1}^{t}\sum_{n=1}^{N}\ytyn{{z}^{(t)}_{nk}}w^{(t)}_n}.
\label{swem}
\end{equation}
Note that we implement Eq. (\ref{swem}) in a recursive manner, \ie, the numerator $\alpha$ and denominator $\beta$ are computed by $\bm{\alpha}^{(t)}_k = \bm{\alpha}^{(t-1)}_k + \sum_{n=1}^{N}\ytyn{{z}^{(t)}_{nk}}w^{(t)}_n\mathbf{x}^{(t)}_n$ and $\bm{\beta}^{(t)}_k = \bm{\beta}^{(t-1)}_k + \sum_{n=1}^{N}\ytyn{{z}^{(t)}_{nk}}w^{(t)}_n$, then $\bm{\mu}^{(t)}_k = \bm{\alpha}^{(t)}_k/\bm{\beta}^{(t)}_k$.

%\begin{equation}
%\begin{aligned}
%    \bm{\alpha}^{(t)}_k &= \bm{\alpha}^{(t-1)}_k + \sum_{n=1}^{N}\mathbf{Z}^{(t)}_{nk}w^{(t)}_n\mathbf{x}^{(t)}_n, \\
%    \bm{\beta}^{(t)}_k &= \bm{\beta}^{(t-1)}_k + \sum_{n=1}^{N}\mathbf{Z}^{(t)}_{nk}w^{(t)}_n.
%\end{aligned}
%\end{equation}
%And then, 
%\begin{equation}
%    \bm{\mu}^{(t)}_k = \bm{\alpha}^{(t)}_k/\bm{\beta}^{(t)}_k.
%\end{equation}
This sequential way of computing base assignment achieves more smooth and adaptable model updating. Instead of storing all frame bases, maintaining only one set of adaptable bases is undoubtedly more friendly to hardware and can also help realize a real-time VOS system. Algorithm\ref{alg:SWEM} shows the detailed pipeline of our SWEM at time step $t$.%

It is also worth noting that the updating of bases is lazy in SWEM.
Since $\ytyn{{z}^{(t)}_{nk}}$ represents the similarity degree between $\mathbf{x}^{(t)}_n$ and $\bm{\mu}^{(t)}_k$, and if a base feature has more similar features with the current frame, it will be updated more quickly. This lazy updating strategy can help SWEM be more robust to noises and prevent drifts. In another way, $w^{(t)}_n$ also enables the hard samples to be updated faster. 

\begin{algorithm}
\caption{The SWEM at the time step $t$}\label{alg:SWEM}
\LinesNumbered
\KwIn{\\
\quad features of frame $t$: \\
\quad \quad $\mathbf{X}^{(t)}\in \mathbb{R}^{N\times C}$, \\
\quad mask of frame $t$: \\
\quad \quad $m^{fg, (t)}\in [0, 1]^{N}$ and $m^{bg, (t)}\in [0, 1]^{N}$ \\
\quad bases at time step $t-1$: \\
\quad \quad $\ytyn{\mathcal{M}}^{fg, (t-1)} \in \mathbb{R}^{K\times C}$ and $\ytyn{\mathcal{M}}^{bg, (t-1)} \in \mathbb{R}^{K\times C}$,  \\
\quad accumulated numerators and denominators: \\
\quad \quad $\bm{\alpha}^{fg, (t-1)},\ \bm{\alpha}^{bg, (t-1)} $ and $\bm{\beta}^{fg, (t-1)}, \bm{\beta}^{bg, (t-1)}$}
\KwOut{\\
\quad bases $\ytyn{\mathcal{M}}^{fg, (t)}$ and $\ytyn{\mathcal{M}}^{bg, (t)}$}
%\quad accumulated responsibilities $\zeta^{fg, (t)}$ and $\zeta^{bg, (t)} \in \mathbb{R}^{K}$}
{\small \tcc{\CUT{The following procedures are performed for both $fg$ and $bg$,} Superscript symbols $fg$ and $bg$ are omitted for simplicity.}}
$\ytyn{\mathcal{M}}^{(t)}\gets \ytyn{\mathcal{M}}^{(t-1)}$ \\ 
$w^{(t)}\gets m^{(t)}$
 %Setting temperature $\tau$ as a constant.

\For{$r=1\textrm{ to } R$}{
	{\small \tcp{SW-\textbf{E} step, estimate responsibilities:}}
     $\ytyn{{z}^{(t)}_{nk}}\gets \frac{\mathcal{K}(\mathbf{x}^{(t)}_{n}, \bm{\mu}^{(t)}_{k})}{\sum_{j=1}^{K}\mathcal{K}(\mathbf{x}^{(t)}_{n},\bm{\mu}^{(t)}_{j})}$ 

	{\small \tcp{SW-\textbf{M} step, update bases:}}
     $\bm{\alpha}^{(t)}_k \gets \bm{\alpha}^{(t-1)}_k + \sum_{n=1}^{N}\ytyn{{z}^{(t)}_{nk}}w^{(t)}_n\mathbf{x}^{(t)}_n$ \\
     $\bm{\beta}^{(t)}_k \gets \bm{\beta}^{(t-1)}_k + \sum_{n=1}^{N}\ytyn{{z}^{(t)}_{nk}}w^{(t)}_n$ \\
     $\bm{\mu}^{(t)}_k \gets \bm{\alpha}^{(t)}_k/\bm{\beta}^{(t)}_k$

	{\small \tcp{SW-\textbf{W} step, calculate weights:}}
	$P^{fg}(\mathbf{x}^{(t)}_n)\gets \frac{\sum_{k=1}^{K} \mathcal{K}(\mathbf{x}^{(t)}_n, \bm{\mu}^{fg, (t)}_k)}{\sum_{k=1}^{K} [\mathcal{K}(\mathbf{x}^{(t)}_n, \bm{\mu}^{fg, (t)}_k)+ \mathcal{K}(\mathbf{x}^{(t)}_n, \bm{\mu}^{bg, (t)}_k)]}$ \\
	$P^{bg}(\mathbf{x}^{(t)}_n) \gets 1-P^{fg}(\mathbf{x}^{(t)}_n)$ \\
	$w^{fg, (t)}_n \gets m^{fg, (t)}_n P^{bg}(\mathbf{x}^{(t)}_n)$ \\
	$w^{bg, (t)}_n \gets m^{bg, (t)}_n P^{fg}(\mathbf{x}^{(t)}_n)$
}
$n=1,2,...,N; \ k=1,2..,K$
\end{algorithm}

\subsection{Matching-based Pipeline}
\label{subsec:pipeline}
The overview of the proposed SWEM network is illustrated in Figure \ref{fig:SWEM}. The whole pipeline mainly consists of three stages, including 1) feature encoding, 2) segmentation, and 3) memorization. 

\noindent{\textbf{Encoding.}}
Similar to previous matching-based methods with Space-Time Memory~\cite{Oh_2019_ICCV,seong_2020_ECCV,li_2020_ECCV_GCM,liang2020video, hu2021learning,xie2021efficient,wang2021swiftnet,seong2021hierarchical,cheng2021stcn}, frames are encoded into \textbf{Key}-\textbf{Value} pairs ($\mathbf{K}\in \mathbb{R}^{N\times C}$ and $\mathbf{V} \in \mathbb{R}^{N\times C'}$) for memory query and read. We adopt the encoder structure of STCN~\cite{cheng2021stcn} to extract image features. The key features are also reused for memorization. Specific network details are described in Section \ref{subsec:network}.

\noindent{\textbf{Segmentation.}}
The segmentation stage includes feature matching and mask decoding. At time step $t$, the $\textbf{Key}$ features $\mathbf{K}^{(t)}$ are used as query to read memory features from $\bm{\kappa}^{(t-1)}$ and $\bm{\nu}^{(t-1)}$, where $\bm{\kappa}$ and $\bm{\nu}$ are base features corresponding to the key and value features and they are concatenations of foreground and background bases ($[\bm{\kappa}^{fg}; \bm{\kappa}^{bg}]\in \mathbb{R}^{2K\times C}$ and $[\bm{\nu}^{fg}; \bm{\nu}^{bg}]\in \mathbb{R}^{2K\times C'}$).
A non-local~\cite{nonlocal} formed matching process is applied as follows:
%\vspace{-0.2cm}
\begin{equation}
 \hat{{\mathbf{V}}}^{(t)}_n = \sum_{k=1}^{2K}\frac{\mathcal{K}({\mathbf{K}}^{(t)}_n, \bm{\kappa}^{(t-1)}_k)}{\sum_{j=1}^{2K}\mathcal{K}({\mathbf{K}}^{(t)}_n, \bm{\kappa}^{(t-1)}_j)} \bm{\nu}^{(t-1)}_k.
 \label{eq:matching_v}
\end{equation}
$\hat{\mathbf{V}}^{(t)}$ is a low-rank reconstruction using memory bases, which is helpful to segmentation tasks. Different from the previous STM-like methods, our memory bases are explicitly separated into foreground and background. Therefore, the correlations $\mathcal{K}(\mathbf{K}^{(t)}_n, \bm{\kappa}^{fg, (t-1)}_k)$ and $\mathcal{K}(\mathbf{K}^{(t)}_n, \bm{\kappa}^{bg, (t-1)}_k)$ can also be used as important segmentation clues. However, the rank of base features is not fixed for different videos since unsorted correlations are not suitable as the inputs for CNNs. To tackle this problem, we design a permutation-invariant operation, which can produce segmentation clues from correlations. Define $\mathcal{K}^{fg, (t)}_n \in \mathbb{R}^{K}$ and $\mathcal{K}^{bg, (t)}_n \in \mathbb{R}^{K}$ as the correlations of $\mathbf{K}^{(t)}_n$ with all foreground and background bases, respectively. The permutation-invariant feature $\mathbf{S}^{(t)}$ can be calculated by:
%\vspace{-0.3cm}
\begin{equation}
\mathbf{S}^{(t)}_{nl} = \frac{\sum_{j\in \text{topl}(\mathcal{K}^{fg, (t)}_n)} \mathcal{K}^{fg, (t)}_{nj}}{\sum_{j\in \text{topl}(\mathcal{K}^{fg, (t)}_n)} \mathcal{K}^{fg, (t)}_{nj} + \sum_{j\in \text{topl}(\mathcal{K}^{bg, (t)}_n)} \mathcal{K}^{bg, (t)}_{nj}},
\label{eq:matching_s}
\end{equation}
where $l=1,2,...,L$. Note that $L \leq K$ is a hyperparameter to control the number of segmentation clues channel and the computation complexity. Besides, $\text{topl}(\cdot)$ means the top-$l$ correlation values.

The decoder takes segmentation clues $[\hat{\mathbf{V}}^{(t)}; \mathbf{S}^{(t)}]$ as input to produce the final mask $\tilde{m}^{(t)}$. Additional skip-connections are also adopted to make use of low-level appearance features.  

%All segmentation clues $[\mathcal{V}^{(t)}; \hat{\mathcal{V}}^{(t)}; \mathbf{S}^{(t)}]\in \mathbb{R}^{N\times 2C'+L}$ are reshaped as $H\times W \times (2C'+L)$ and received by the decoder to produce the final mask $\tilde{m}^{(t)}$. Additional skip-connections are also adopted to make use of low-level appearance features.   

\noindent{\textbf{Memorization.}}
After the segmentation, key features $\mathbf{K}^{(t)}$ are reused for the memorization stage. We adopt another ResNet-18 to re-encode the image-mask pair to obtain value features $\mathbf{V}^{(t)}$. The key bases are updated by $\bm{\kappa}^{(t)}=\rm{SWEM}(\mathbf{K}^{(t)}, \tilde{m}^{(t)}, \bm{\kappa}^{(t-1)})$ which is described in Algorithm \ref{alg:SWEM}. To maintain the alignment between key and value, the updated value bases is calcudated by $\bm{\nu}^{(t)}_k = (\bm{\beta}^{(t-1)}_k \bm{\nu}^{(t-1)}_k + \sum_{n=1}^{N}\ytyn{{z}^{(t)}_{nk}}w^{(t)}_n\mathbf{v}^{(t)}_n)/\bm{\beta}^{(t)}_k$, where $\bm{\beta}$, $\mathbf{Z}$ and $w$ are all produced during the construction of key bases $\bm{\kappa}$.

%%%%%%%%%%%%%%%%%%%%%%%%%%%%%%%%%%%%%%%%%%%%%%%%%%%%%%%%%%%%%%%%%%%%%%%%%%%%%%%%%%%%%%%%%%%%%%%%%%%%%%%%%%%%%%%%%%%%%%% Implement Details
\section{Implementation Details}

\subsection{Network Structure}
\label{subsec:network}
We adopt ResNet-50~\cite{he2016deep} as the backbone to extract frame features and ResNet-18 for value feature extraction. All batch normalization layers are frozen. The stage 4 (res4) features are used for memorization and matching processes. These feature maps have a stride of 16 compared with the row image. The temperature hyper-parameter $\tau$ is set to 0.05. The number of base features in a group is set as $K=128$, and the number of iterations $R$ is set as 4 in the SWEM algorithm. We select top-64 ($L=64$) correlation scores calculated by Eq.~\eqref{eq:matching_s}. For simplicity and fair comparison with STM~\cite{Oh_2019_ICCV}, we use the same two-level decoder, which consists of two refining layers, and each layer contains two residual blocks. 

%The final output is bounded by a SoftMax function and resized to align the raw input.

\subsection{Two-stage Training}
\noindent{\textbf{Pre-training on static image datasets.}}
Following the previous methods~\cite{Wang_2019_ICCV,Oh_2019_ICCV,seong_2020_ECCV,lu_2020_ECCV,liang2020video}, we first perform the pre-training procedure on static image datasets~\cite{cocolin2014microsoft,msra10kChengPAMI,ecssdshi2015hierarchical,pascalsli2014secrets,pascolvoceveringham2010pascal}. The input frames are cropped into $384\times 384$ for training. Three frames are generated based on a single image at each step, where the random affine transformations of shearing, rotation, scaling, and cropping are applied. The Adam optimizer~\cite{kingma2014adam} with the learning rate $1e$-$5$ is adopted for all training processes. Besides, we use the cross-entropy loss for the final segmentation.

\noindent{\textbf{Training on video datasets.}}
After pre-training on images, we fine-tune the proposed SWEM on video datasets DAVIS 2017~\cite{pont20172017} and the YouTube-VOS 2018~\cite{xu2018youtube}. The training process is similar to image pre-training, where the main difference is that we sample the three frames from a video clip randomly instead of one single image. For multi-object frames, we randomly select less than 3 objects. We perform all experiments on a single NVIDIA Tesla V100 GPU with a batch size of 4. \CUT{and we also evaluate our approach on an NVIDIA 1080ti GPU.}

%%%%%%%%%%%%%%%%%%%%%%%%%%%%%%%%%%%%%%%%%%%%%%%%%%%%%%%%%%%%%%%%%%%%%%%%%%%%%%%%%%%%%%%%%%%%%%%%%%%%%%%%%%%%%%%%%%%%%%% Experiments
\section{Experiments}

%%%%%%%%%%%%%%%%%%%%%%%%%%%%%%%%%%%%%%%%%%%%%%%%%%%%%%%%%%%%%%%%%%%% Table Ablation of number of memories
\begin{table}[t]
\begin{center}
\begin{tabular}{ccccccc}
\toprule 
 \multirow{2}*{$K$} & \multirow{2}*{FPS}& \multicolumn{2}{c}{DAVIS 2016 val} & \multicolumn{2}{c}{DAVIS 2017 val} \\
%\cline{5-10}
 &    & $\mathcal{J}$ \& $\mathcal{F}$ $\uparrow$ & $\mathcal{J}_M$ $\uparrow$ & $\mathcal{J}$ \& $\mathcal{F}$ $\uparrow$ & $\mathcal{J}_M$ $\uparrow$ \\
\midrule 
   32   & 37.3 & 88.4 & 87.6 & 80.2 & 77.7 \\
   64   & 36.8 & 88.9 & 88.0 & 80.9 & 78.4 \\
   128  & 36.4 & 89.5 & 88.6 & 81.9 & 79.3 \\
   256  & 35.5 & 89.5 & 88.5 & 82.0 & 79.4 \\
\bottomrule
\end{tabular}
\vspace{-0.4cm}
\end{center}
\caption{Ablation study on the number of bases $K$ (with $R$=4).}
\label{tab:ablation_k}
\vspace{-0.4cm}
\end{table}
%%%%%%%%%%%%%%%%%%%%%%%%%%%%%%%%%%%%%%%%%%%%%%%%%%%%%%%%%%%%%%%%%%%% Table Ablation of number of memories

%%%%%%%%%%%%%%%%%%%%%%%%%%%%%%%%%%%%%%%%%%%%%%%%%%%%%%%%%%%%%%%%%%%% Table SOTAs on DAVIS16 and DAVIS17
\begin{table*}[!t]
\begin{center}
\begin{tabular}{cccccccccc}
\toprule 
\multirow{2}*{Method} & \multirow{2}*{Pub.} & \multirow{2}*{I} & \multirow{2}*{FPS} & \multicolumn{3}{c}{DAVIS 2016 val} & \multicolumn{3}{c}{DAVIS 2017 val} \\
%\cline{5-10}
 &  &  &  & $\mathcal{J}$ \& $\mathcal{F}$ $\uparrow$ & $\mathcal{J}_M$ $\uparrow$ & $\mathcal{F}_M$ $\uparrow$ & $\mathcal{J}$ \& $\mathcal{F}$ $\uparrow$ & $\mathcal{J}_M$ $\uparrow$ & $\mathcal{F}_M$ $\uparrow$ \\
\midrule
%FTRM~\cite{Robinson_2020_CVPR}                      & CVPR 2020          &            & 22        & 81.7 & -    & -    & 68.8 & -    & -    \\
%LWL~\cite{Bhat_2020_ECCV_GCM}                       & ECCV 2020          &            & 6         & -    & -    & -    & 74.3 & 72.2 & 76.3 \\
STM~\cite{Oh_2019_ICCV}                             & ICCV 2019          & \checkmark & 6         & 86.5 & 84.8 & 88.1 & 71.6 & 69.2 & 74.0 \\
AFB-URR~\cite{liang2020video}                       & NeuralPS 2020      & \checkmark & 4         & -    & -    & -    & 74.6 & 73.0 & 76.1 \\
CFBI~\cite{yang_2020_ECCV}                          & ECCV 2020          &            & 5         & 86.1 & 85.3 & 86.9 & 74.9 & 72.1 & 77.7 \\
%KMN~\cite{seong_2020_ECCV}                          & ECCV 2020          & \checkmark & 8         & \underline{87.6} & 87.1 & \underline{88.1} & \underline{76.0} & \underline{74.2} & 77.8 \\
\midrule
%SiamMask~\cite{wang2019fast}                        & ResNet-50          &            & 36        & 70.0 & 71.7 & 67.8 & 56.4 & 54.3 & 58.5 \\
RANet~\cite{Wang_2019_ICCV}                         & ICCV 2019          & \checkmark & 30        & 85.5 & 85.5 & 85.4 & 65.7 & 63.2 & 68.2 \\
GC~\cite{li_2020_ECCV_GCM}                          & ECCV 2020          & \checkmark & 25        & 86.6 & \textbf{87.6} & 85.7 & 71.4 & 69.3 & 73.5 \\
TVOS~\cite{Zhang_2020_CVPR}                         & CVPR 2020          &            & 37        & -    & -    & -    & 72.3 & 69.9 & 74.7 \\
SAT~\cite{Chen_2020_CVPR}                           & CVPR 2020          &            & 39        & 83.1 & 82.6 & 83.6 & 72.3 & 68.6 & 76.0 \\
\midrule
% NeuralPS 2021
\textbf{SWEM}     &  CVPR2022    &            & 36        & \textbf{88.1} & 87.3 & \textbf{89.0} & \textbf{77.2} & \textbf{74.5} & \textbf{79.8} \\
%\textbf{SWEM}                                     & ResNeXt-50         &            & 27        & \textbf{88.7} & 87.5 & \textbf{89.9} & \textbf{77.9} & \textbf{75.2} & 80.7 \\
\midrule
\midrule
%FTRM~\cite{Robinson_2020_CVPR}(+\textbf{YV})        & CVPR 2020          &            & 22        & 83.5 & -    & -    & 76.7 & -    & -    \\
%LWL~\cite{Bhat_2020_ECCV_GCM}(+\textbf{YV})         & ECCV 2020          & \checkmark & 6         & -    & -    & -    & 81.6 & 79.1 & 84.1 \\
%FEELVOS~\cite{Voigtlaender_2019_CVPR}(+\textbf{YV}) & CVPR 2019          &            & 2         & 81.7 & 81.1 & 82.2 & 71.5 & 69.1 & 74.0 \\
STM~\cite{Oh_2019_ICCV}(+\textbf{YV})               & ICCV 2019          & \checkmark & 11*        & 89.3 & 88.7 & 89.9 & 81.7 & 79.2 & 84.3 \\
CFBI~\cite{yang_2020_ECCV}(+\textbf{YV})            & ECCV 2020          & \checkmark & 5         & 89.4 & 88.3 & 90.5 & 81.9 & 79.1 & 84.6 \\
EGMN~\cite{lu_2020_ECCV}(+\textbf{YV})              & ECCV 2020          & \checkmark & 5         & -    & -    & -    & 82.8 & 80.2 & 85.2 \\
KMN~\cite{seong_2020_ECCV}(+\textbf{YV})            & ECCV 2020          & \checkmark & 8         & 90.5 & 89.5 & 91.5 & 82.8 & 80.0 & 85.6 \\
% CVPR 2021
SSTVOS~\cite{duke2021sstvos} (+\textbf{YV})         & CVPR 2021          &            & $\sim$ 7  & - & - & - & 82.5 & 79.9 & 85.1 \\
RMNet~\cite{xie2021efficient} (+\textbf{YV})          & CVPR 2021          & \checkmark & 12        & 88.8 & 88.9 & 88.7 & 83.5 & 81.0 & 86.0 \\
LCM~\cite{hu2021learning}   (+\textbf{YV})          & CVPR 2021          & \checkmark & 9         & 90.7 & 89.9 & 91.4 & 83.5 & 80.5 & 86.5 \\

JOINT~\cite{mao2021joint}   (+\textbf{YV})          & ICCV 2021          &            & 4         & - & - & - & 83.5 & 80.8 & 86.2 \\
DMN~\cite{liang2021video}   (+\textbf{YV})          & ICCV 2021          & \checkmark & 7         & - & - & - & 84.0 & 81.0 & 87.0 \\
HMMN~\cite{seong2021hierarchical} (+\textbf{YV})    & ICCV 2021          & \checkmark & 10        & 90.8 & 89.6 & 92.0 & 84.7 & 81.9 & 87.5 \\
AOT~\cite{yang2021associating} (+\textbf{YV})       & NeuralPS 2021      & \checkmark & 19        & 91.0 & 89.7 & 92.3 & 83.0 & 80.3 & 85.7 \\
\midrule
Swift~\cite{wang2021swiftnet} (+\textbf{YV})        & CVPR 2021          & \checkmark & 25        & 90.4 & 90.5 & 90.3 & 81.1 & 78.3 & 83.9 \\
STCN~\cite{cheng2021stcn} (+\textbf{YV})            & NeuralPS 2021      & \checkmark & 26*       & \textbf{91.6} & \textbf{90.8} & 92.5 & \textbf{85.4} & \textbf{82.2} & \textbf{88.6} \\
\midrule
\textbf{SWEM}(+\textbf{YV})  &  CVPR2022 &  \checkmark & 36        & 91.3 & 89.9 & \textbf{92.6} & 84.3 & 81.2 & 87.4 \\
%\textbf{SWEM}(+\textbf{YV})                       & ResNeXt-50         &            & 27        & -    & -    & -    & -    & -    & -    \\
\bottomrule

\end{tabular}
\end{center}
\vspace{-0.4cm}
\caption{Comparisons with previous approaches on DAVIS 2016 and DAVIS 2017 validation sets. `\textbf{+YV}' denotes training with additional videos from YouTube-VOS. `I' indicates the pre-training on image datasets. Note that our SWEM achieves results close to state-of-the-art performance at a speed of 36 FPS on a V100 GPU without IO time. Here, '*' represents the re-evaluation on our hardware for reference.}
\label{tab:DAVIS_quantity}
\vspace{-0.4cm}
\end{table*}
%%%%%%%%%%%%%%%%%%%%%%%%%%%%%%%%%%%%%%%%%%%%%%%%%%%%%%%%%%%%%%%%%%%% Table SOTAs on DAVIS16 and DAVIS17

\subsection{Ablation Study}
We first analyze the impact of the number of bases $K$ and that of SWEM iterations $R$, which are key factors affecting the efficiency of the model. Then we investigate the effect of adaptive weights in SWEM on model performance. We directly train all models on video datasets without pre-training on images. Models are evaluated on DAVIS 2016~\cite{perazzi2016benchmark} and DAVIS 2017 validation datasets.   

% base的个数，iteration的数目
\noindent{\textbf{The number of bases $K$.}} %The number of bases $K$ is set as 32, 64, 128, and 256 separately. 
Table~\ref{tab:ablation_k} shows the quantitative results and inference speed under different $K$ values. The performance saturates at $K=128$. When decreasing the number of bases, the performance degraded a lot, while this does not save too much computation, as seen in the inference speed. Therefore, we choose a relatively large $K=128$ as the default setting.

% 表达迭代次数对速度的影响非常大，我们在效率和性能上面做权衡，选择了4。
\noindent{\textbf{The number of SWEM iterations $R$.}} The number of SWEM iterations affects the efficiency and convergence of bases construction. Table~\ref{tab:ablation_r} shows results with $R=1\sim 7$. The inference speed is sensitive to the number of iterations. Every increase in $R$ decreases the inference speed by $1\sim 2$ FPS. $R=4$ achieves the best trade-off between performance and efficiency.

%%%%%%%%%%%%%%%%%%%%%%%%%%%%%%%%%%%%%%%%%%%%%%%%%%%%%%%%%%%%%%%%%%%% Table Ablation of number of SWEM iterations
\begin{table}[b]
\begin{center}
\begin{tabular}{ccccccc}
\toprule 
 \multirow{2}*{$R$} & \multirow{2}*{FPS}& \multicolumn{2}{c}{DAVIS 2016 val} & \multicolumn{2}{c}{DAVIS 2017 val} \\
%\cline{5-10}
 &    & $\mathcal{J}$ \& $\mathcal{F}$ $\uparrow$ & $\mathcal{J}_M$ $\uparrow$ & $\mathcal{J}$ \& $\mathcal{F}$ $\uparrow$ & $\mathcal{J}_M$ $\uparrow$ \\
\midrule 
   1   & 41.5 & 87.7 & 87.3 & 77.9 & 75.1 \\
   2   & 39.4 & 88.7 & 88.0 & 79.5 & 76.9 \\
   3   & 38.3 & 88.8 & 87.9 & 80.8 & 78.1 \\
   4   & 36.4 & 89.5 & 88.6 & 81.9 & 79.3 \\
   5   & 34.5 & 89.1 & 88.2 & 81.2 & 78.4 \\
   6   & 33.0 & 89.0 & 88.3 & 79.8 & 77.0 \\
   7   & 31.8 & 88.6 & 87.8 & 79.8 & 77.1 \\
\bottomrule
\end{tabular}
\vspace{-0.4 cm}
\end{center}
\caption{Ablation study on the number of SWEM iterations $R$ (with $K$=128).}
\label{tab:ablation_r}
\vspace{-0.5 cm}
\end{table}
%%%%%%%%%%%%%%%%%%%%%%%%%%%%%%%%%%%%%%%%%%%%%%%%%%%%%%%%%%%%%%%%%%%% Table Ablation of number of SWEM iterations

% adaptive weights 的有效性

\begin{figure}[b]
\begin{center}
\includegraphics[width=1.0\linewidth]{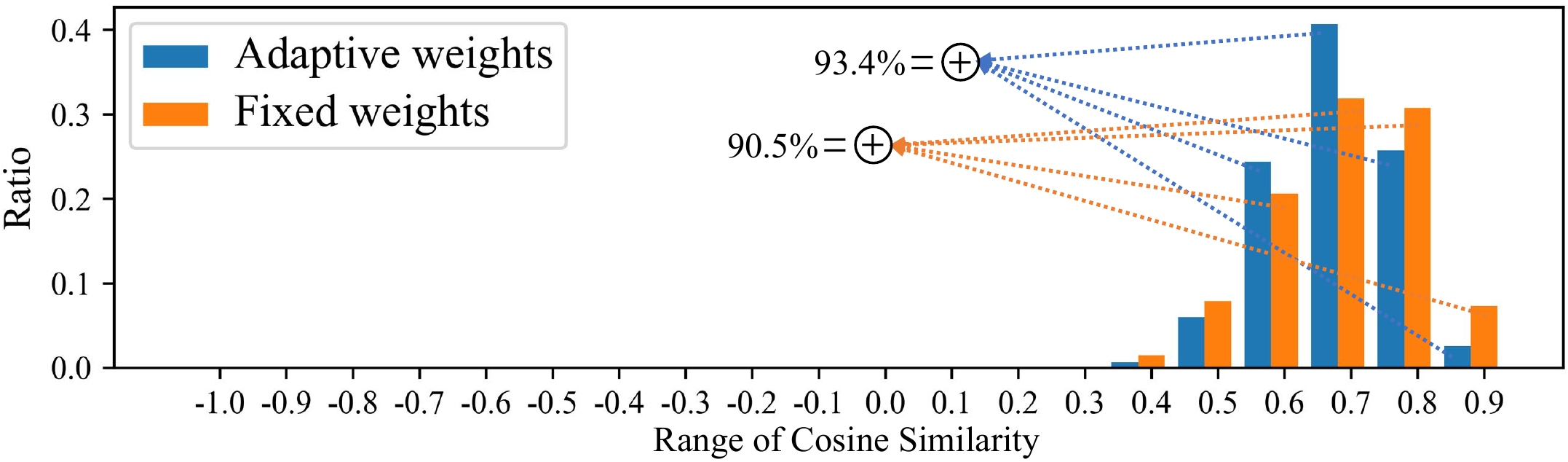}
\end{center}
\vspace{-0.4 cm}
   \caption{The distribution of maximum matching similarities between current frame features and previous bases.}
\label{fig:ablation_w}
\vspace{-0.5 cm}
\end{figure}

\begin{figure}[t]
\begin{center}
\includegraphics[width=1\linewidth]{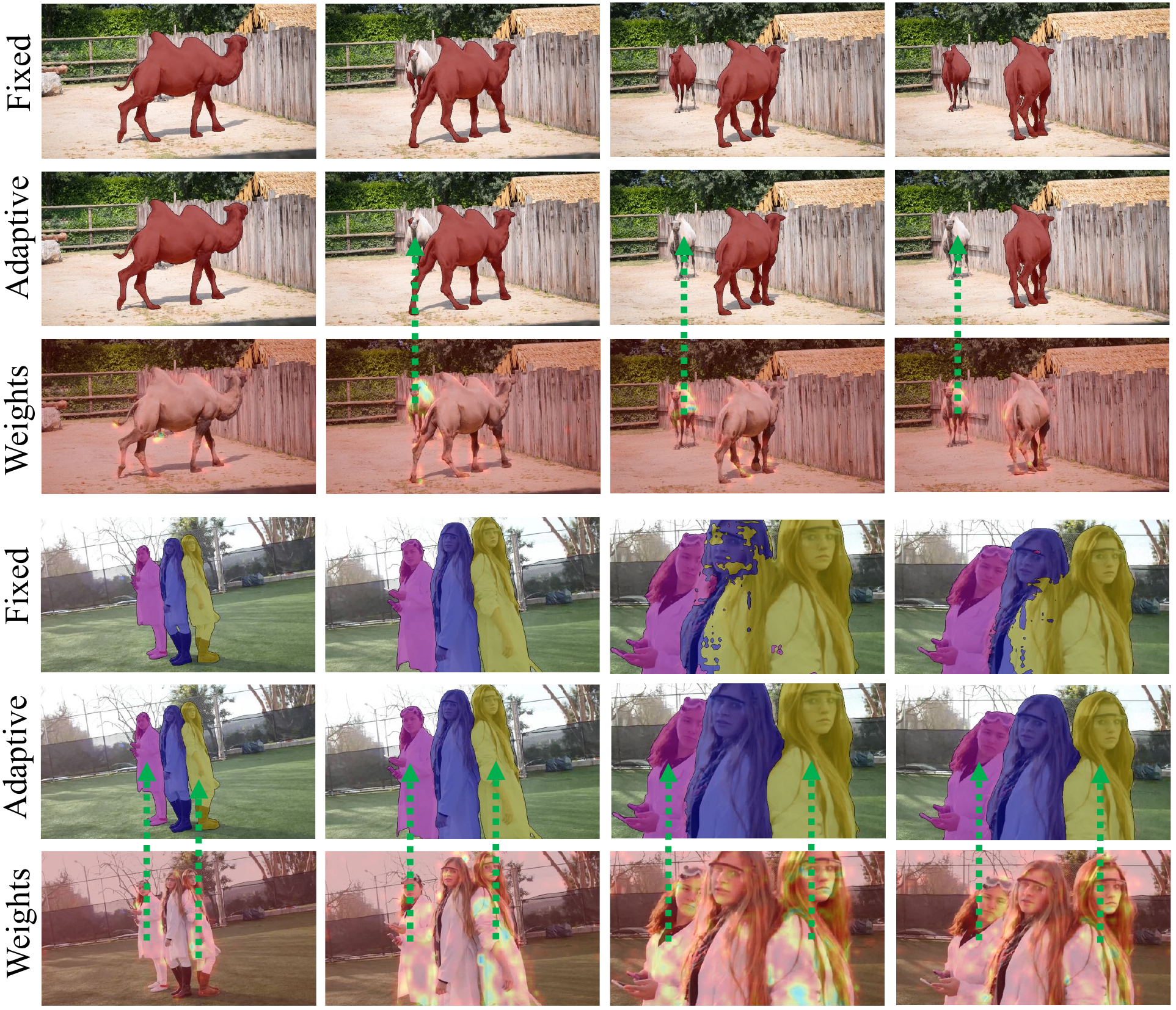}
\end{center}
\vspace{-0.4 cm}
   \caption{The qualitative comparison between adaptive-weights and fixed-weights. SWEM with fixed weights (first line for each sample) is struggled to distinguish similar objects while {the one with} adaptive weights (second line) {is} competent for this problem. Weights are visualized at the third line for each sample, where the brighter the pixel, the harder it is. As for the person in the middle, the corresponding background bases pay more attention to objects (the remaining person) which are similar to the target.}
\label{fig:ablation_wf}
\vspace{-0.5 cm}
\end{figure}

\noindent{\textbf{Adaptive weights in SWEM.}} 
Without using adaptive weights (Eq.~\ref{eq:adapt_w}), our performance drops greatly (81.9\% $\rightarrow$ 77.6\%) while the improvement of inference speed is subtle (36.4 FPS $\rightarrow$ 38.4 FPS). 
Figure~\ref{fig:ablation_w} shows the distribution of maximum matching similarities between features of the current frame and previous bases. Although SWEM with adaptive weights has fewer high similarities, it has more similarities above 0.6 than the one with fixed weights (93.4\% v.s. 90.5\%), guaranteeing fewer missing matches during inference. 
We also show the qualitative comparison between two kinds of weights in Figure~\ref{fig:ablation_wf}. Compared with SWEM with adaptive weights, that with fixed weights is more prone to missing matches, resulting in drift issues.

%%%%%%%%%%%%%%%%%%%%%%%%%%%%%%%%%%%%%%%%%%%%%%%%%%%%%%%%%%%%%%%%%%%% Figure Ablation of adaptive weights
%%%%%%%%%%%%%%%%%%%%%%%%%%%%%%%%%%%%%%%%%%%%%%%%%%%%%%%%%%%%%%%%%%%% Figure Ablation of adaptive weights

%%%%%%%%%%%%%%%%%%%%%%%%%%%%%%%%%%%%%%%%%%%%%%%%%%%%%%%%%%%%%%%%%%%% Figure Ablation of adaptive weights
%%%%%%%%%%%%%%%%%%%%%%%%%%%%%%%%%%%%%%%%%%%%%%%%%%%%%%%%%%%%%%%%%%%% Figure Ablation of adaptive weights

%%%%%%%%%%%%%%%%%%%%%%%%%%%%%%%%%%%%%%%%%%%%%%%%%%%%%%%%%%%%%%%%%%%%%%%%%%%%%%%%%%%%%%%%%%%%%%%%%%%%%% Comparison with SOTA
\subsection{Comparison with SOTA}

%%%%%%%%%%%%%%%%%%%%%%%%%%%%%%%%%%%%%%%%%%%%%%%%%%%%%%%%%%%%%%%%%%%% Table SOTAs on YTVOS18
\begin{table}[t]
\begin{center}
\begin{tabular}{cccccc}
\toprule 
  \multirow{2}*{Method} &\multirow{2}*{$\mathcal{G}$} & \multicolumn{2}{c}{seen} & \multicolumn{2}{c}{unseen} \\
%\cline{3-6}
 & & $\mathcal{J}_M$ $\uparrow$& $\mathcal{F}_M$ $\uparrow$ & $\mathcal{J}_M$ $\uparrow$ & $\mathcal{F}_M$ $\uparrow$\\
\midrule

%AGAME~\cite{Johnander_2019_CVPR}       & 66.1 & 67.8 &  -   & 60.8 &  -   \\
%PreMVOS~\cite{luiten2018premvos}       & 66.9 & 71.4 & 75.9 & 56.5 & 63.7 \\
STM~\cite{Oh_2019_ICCV}                & 79.4 & 79.7 & 84.2 & 72.8 & 80.9 \\
AFB-URR~\cite{liang2020video}          & 79.6 & 78.8 & 83.1 & 74.1 & 82.6 \\
EGMN~\cite{lu_2020_ECCV}               & 80.2 & 80.7 & 85.1 & 74.0 & 80.9 \\
KMN~\cite{seong_2020_ECCV}             & 81.4 & 81.4 & 85.6 & 75.3 & 83.3 \\
CFBI~\cite{yang_2020_ECCV}             & 81.4 & 81.1 & 85.8 & 75.3 & 83.4 \\
%LWL~\cite{Bhat_2020_ECCV_GCM}          & 81.5 & 80.4 & 84.9 & 76.4 & 84.4 \\
RMNet~\cite{xie2021efficient}            & 81.5 & 82.1 & 85.7 & 75.7 & 82.4 \\
SSTVOS~\cite{duke2021sstvos}           & 81.7 & 81.2 & 85.9 & 76.0 & 83.9 \\
LCM~\cite{hu2021learning}              & 82.0 & 82.2 & 86.7 & 75.7 & 83.4 \\
DMN~\cite{liang2021video}              & 82.5 & 82.5 & 86.9 & 76.2 & 84.2 \\
HMMN~\cite{seong2021hierarchical}      & 82.6 & 82.1 & 87.0 & 76.8 & 84.6 \\
JOINT~\cite{mao2021joint}              & 83.1 & 81.5 & 85.9 & \textbf{78.7} & 86.5 \\
AOT~\cite{yang2021associating}         & \textbf{83.7} & \textbf{82.5} & \textbf{87.5} & 77.9 & \textbf{86.7} \\
\midrule
SAT*~\cite{Chen_2020_CVPR}             & 63.6 & 67.1 & 70.2 & 55.3 & 61.7 \\
TVOS*~\cite{Zhang_2020_CVPR}           & 67.8 & 67.1 & 69.4 & 63.0 & 71.6 \\
FRTM*~\cite{Robinson_2020_CVPR}        & 72.1 & 72.3 & 76.2 & 65.9 & 74.1 \\
GC*~\cite{li_2020_ECCV_GCM}            & 73.2 & 72.6 & 75.6 & 68.9 & 75.7 \\
Swift*~\cite{Bhat_2020_ECCV_GCM}       & 77.8 & 77.8 & 81.8 & 72.3 & 79.5 \\
STCN*~\cite{cheng2021stcn}             & \textbf{83.0} & 81.9 & 86.5 & \textbf{77.9} & \textbf{85.7} \\
\midrule
\textbf{SWEM*}                         & 82.8 & \textbf{82.4} & \textbf{86.9} & 77.1 & 85.0 \\
\bottomrule
\end{tabular}
\end{center}
\vspace{-0.4 cm}
\caption{Comparison with state-of-the-art methods on the YouTube-VOS 2018 validation dataset. We report all of the mean Jaccard ($\mathcal{J}$), the boundary ($\mathcal{F}$) scores for seen and unseen categories, and the overall scores $\mathcal{G}$. Besides, we use `*' to indicate those methods with an inference speed $>20$ FPS. Note SSTVOS, JOINT and AOT are transformer-based methods.}
\label{tab:YTVOS}
\vspace{-0.5 cm}
\end{table}
%%%%%%%%%%%%%%%%%%%%%%%%%%%%%%%%%%%%%%%%%%%%%%%%%%%%%%%%%%%%%%%%%%%% Table SOTAs on YTVOS18

\noindent{\textbf{Datasets and evaluation metrics.}} We report the results on the DAVIS 2016, DAVIS 2017, and YouTube-VOS 2018 datasets using region similarity $\mathcal{J}$, contour accuracy $\mathcal{F}$, and their mean as metrics.

\noindent{\textbf{DAVIS 2016 and DAVIS 2017.}} Table~\ref{tab:DAVIS_quantity} presents the quantitative comparisons with recent state-of-the-art video segmentation methods on the DAVIS 2016 and 2017 validation sets. Our method achieves the best $\mathcal{J}$ \& $\mathcal{F}$ on both datasets among methods without pre-training on image datasets or additional video datasets. In detail, our method outperforms SAT~\cite{Chen_2020_CVPR}, which runs at a similar speed (39 FPS) with ours, with a large margin on DAVIS2017 (+4.9\% $\mathcal{J}$ \& $\mathcal{F}$ score). Under the setting of using additional training data from YouTube-VOS (\textbf{+YV}), our SWEM surpasses all other top-performing methods. Note that STCN~\cite{cheng2021stcn} uses a growing memory bank during inference, which harms the long-term segmentation while SWEM keeps the fixed number of bases and has stable computation complexity. SSTVOS~\cite{duke2021sstvos}, AOT~\cite{yang2021associating} and JOINT~\cite{mao2021joint} adopt transformer~\cite{vaswani2017attention} backbones which are more powerful than ResNet-50. Hierarchical matching used in HMMN~\cite{seong2021hierarchical} also affects the segmentation efficiency deeply. We re-evaluated STM and STCN on our hardware and software environment for reference. Our SWEM is capable of achieving an inference speed of 36 FPS on the V100 GPU and 27 FPS on the 1080ti GPU. 

\noindent{\textbf{YouTube-VOS 2018.}} We make a comparison between our SWEM with previous methods on the YouTube-VOS 2018 via the official evaluation server in Table~\ref{tab:YTVOS}. Note that although SWEM leverages the original ResNet-50 backbone and the identical decoder as STM~\cite{Oh_2019_ICCV}, it achieves the 82.8\% overall score which is very close to the state-of-the-art results. Besides, we provide more qualitative and quantitative comparisons in the Supplementary Material.

%%%%%%%%%%%%%%%%%%%%%%%%%%%%%%%%%%%%%%%%%%%%%%%%%%%%%%%%%%%%%%%%%%%%%%%%%%%%%%%%%%%%%%%%%%%%%%%%%%%%%%%%%%%%%%%%%%%%%%% Conclusion
\section{Conclusion}
In this paper, we proposed a fast yet robust model for semi-supervised video object segmentation dubbed Sequential Weighted Expectation-Maximum (SWEM) network, which is capable of constructing compact target templates with low redundancy for pixel-wise matching. The weighted EM algorithm is used to construct bases for foreground and background features separately and reduce the intra-frame redundancy. We also proposed to compute adaptive weights instead of fixed weights when generating bases, which forces bases to pay more attention to hard samples, so as to reduce the missing match. We extended the weighted EM to sequential weighted EM to process sequential data and completely reduce the inter-frame redundancy. Overall, our method achieves a performance close to the state-of-the-art on VOS at 36 FPS.

\section*{Acknowledgement}
\vspace{-0.0 cm}
This work was supported by NSFC project Grant No.U1833101, SZSTI Grant No.JCYJ20190809172201639 and WDZC20200820200655001, the Joint Research Center of Tencent and Tsinghua.

%%%%%%%%% REFERENCES
{\small
\bibliographystyle{ieee_fullname}
\bibliography{egbib}
}

\clearpage
\appendix
\renewcommand{\appendixname}{Appendix~\Alph{section}}
%\begin{appendices}
\section{Overview}
We provide the supplementary results for our main paper, which are organized as follows. Section~\ref{sec:TraingDetails} introduces more training details not mentioned in the original paper. In Section~\ref{sec:ablation}, we provide more ablation analysis on the number of permutation-invariant features $L$ (i.e.,  Eq.(12) in the main paper) and the temperature hyperparameter $\tau$ in the similarity calculation (i.e. Eq.~\ref{eq:matching_tau} in this supplementary material ). In Section~\ref{sec:eff}, we provide the comparison between the proposed SWEM and STCN~\cite{cheng2021stcn} on both efficiency and performance with a commonly used GPU, NVIDIA GTX 1080ti. We also report SWEM performance on YouTube-VOS 2019 validation set in Section~\ref{sec:YTVOS19}. In Section~\ref{sec:TrainingData}, we analyze the effect of training with different datasets. Finally, we provide qualitative results in Section~\ref{sec:qualitative}.

\section{More Training Details}
\label{sec:TraingDetails}
We set channel dimension as $C=128$ for the \textbf{Key}  and $C'=512$ for the \textbf{Value} feature, which are the same as those in STM~\cite{Oh_2019_ICCV}. 
We first train SWEM on image datasets for 500k iterations, including COCO~\cite{cocolin2014microsoft}, MSRA10K~\cite{msra10kChengPAMI}, ECSSD~\cite{ecssdshi2015hierarchical}, PASCOL-S~\cite{pascalsli2014secrets} and PASCOL-VOC2012~\cite{pascolvoceveringham2010pascal}, and then on video datasets for 200k iterations, including DAVIS~\cite{pont20172017}, and the YouTube-VOS~\cite{xu2018youtube}. When using a single NVIDIA V100 GPU, the image training process takes about 2 days while the video training process needs about 1 day.
Following STM and STCN~\cite{cheng2021stcn}, 
we sample video clips using an 11-times (3471 v.s 300) higher probability for YouTube-VOS than DAVIS 2017 training set. Our SWEM takes about 8GB GPU memory with batch size 4, which reveals its memory-efficient during training and testing.

%%%%%%%%%%%%%%%%%%%%%%%%%%%%%%%%%%%%%%%%%%%%%%%%%%%%%%%%%%%%%%%%%%%%%%%%%%%%%%%%%%%%%%%%%%%%%%%%%%%%%%%%%%%%%%%%%%%%%%% Re
\begin{algorithm}
\caption{Calculation of Permutation-Invariant Affinity Features}\label{alg:S}
\LinesNumbered
\KwIn{$\mathcal{K}^{fg}$:$K \times HW$, $\mathcal{K}^{bg}$:$K \times HW$ }
\KwOut{$\mathbf{S}$:$L \times HW$}

%{\mono test }
\textcolor{blue}{function} GetPermutInvariantFeatures($\mathcal{K}^{fg}$, $\mathcal{K}^{bg}$): \
\quad \textcolor{cyan}{\tcp{Get top-L affinities for each pixel.}}
\quad $\tilde{\mathcal{K}}^{fg/bg}$=torch.topk($\mathcal{K}^{fg/bg}$, k=$L$, dim=$1$) \\
\quad \textcolor{cyan}{\tcp{Initialize $\mathbf{S}^{fg}$ and $\mathbf{S}^{bg}$  as zeros.} }
\quad $\mathbf{S}^{fg/bg}$=torch.zeros\_like($\tilde{\mathcal{K}}^{fg/bg}$) \\
\quad $\mathbf{S}^{fg/bg}_{0}$=$\tilde{\mathcal{K}}^{fg/bg}_{0}$ \\
\quad for $l$ in range($1$, $L$): \\
\quad \quad $\mathbf{S}^{fg/bg}_l$=$\mathbf{S}^{fg/bg}_{l-1}$+$\tilde{\mathcal{K}}^{fg/bg}_{l}$ \\
\quad $\mathbf{S}$=$\mathbf{S}^{fg}$/($\mathbf{S}^{fg}$+$\mathbf{S}^{bg}$) \\
\textcolor{blue}{return} $\mathbf{S}$ 
\end{algorithm}

\section{More ablation analysis}
\label{sec:ablation}

\noindent{\textbf{The number of permutation-invariant features $L$.}} Recall that we introduced a permutation-invariant operation to take advantage of affinities between pixels in the current frame and memory features in Section 4.3 of the main paper. The calculation of affinity features is detailed as Eq.(12) in the main paper, which is repeated as:
\begin{equation}
\mathbf{S}^{(t)}_{nl} = \frac{\sum_{j\in \text{topl}(\mathcal{K}^{fg, (t)}_n)} \mathcal{K}^{fg, (t)}_{nj}}{\sum_{j\in \text{topl}(\mathcal{K}^{fg, (t)}_n)} \mathcal{K}^{fg, (t)}_{nj} + \sum_{j\in \text{topl}(\mathcal{K}^{bg, (t)}_n)} \mathcal{K}^{bg, (t)}_{nj}},
\label{eq:matching_s}
\end{equation}
where $l=1,2,...,L$, $L \leq K$, and $K$ is the number of foreground or background base features.

Define $\mathcal{K}^{fg}$ and $\mathcal{K}^{bg}$ are affinity matrices between frame features and foreground-background base features. Then the permutation-invariant features with $L$ channels can be calculated with Algorithm~\ref{alg:S} in PyTorch~\cite{paszke2019pytorch} style. $L$ is a hyperparameter to control the number of segmentation clues channels and the computation complexity. Table~\ref{tab:ablation_l} shows the results when setting $L$ as 8, 16, 32, 64 and 128 separately. The segmentation performance is robust to various $L$ except $L=8$, which brings a  significant performance reduction. When increasing $L$ from 64 to 128, the inference speed is dropped by 3.4 FPS while the performance is not improved. 
\begin{table}[t]
\begin{center}
\begin{tabular}{ccccccc}
\toprule 
 \multirow{2}*{$L$} & \multirow{2}*{FPS}& \multicolumn{2}{c}{DAVIS 2016 val} & \multicolumn{2}{c}{DAVIS 2017 val} \\
%\cline{5-10}
 &    & $\mathcal{J}$ \& $\mathcal{F}$ $\uparrow$ & $\mathcal{J}_M$ $\uparrow$ & $\mathcal{J}$ \& $\mathcal{F}$ $\uparrow$ & $\mathcal{J}_M$ $\uparrow$ \\
\midrule 
   8   & 40.8 & 88.5 & 87.5 & 81.0 & 78.6 \\
   16   & 39.1 & 89.5 & 88.6 & 81.4 & 78.8 \\
   32  & 37.5 & \textbf{89.6} & \textbf{88.8} & 81.5 & 78.8 \\
   64  & 36.4 & 89.5 & 88.6 & \textbf{81.9} & \textbf{79.3} \\
   128  & 33.0 & 89.6 & 88.6 & 81.6 & 79.0 \\
\bottomrule
\end{tabular}
\vspace{-0.0 cm}
\end{center}
\caption{Ablation study on the number of permutation-invariant features $L$.}
\label{tab:ablation_l}
\vspace{-0.3 cm}
\end{table}

\begin{table}[b]
\begin{center}
\begin{tabular}{cccccc}
\toprule 
 \multirow{2}*{$\tau$} & \multicolumn{2}{c}{DAVIS 2016 val} & \multicolumn{2}{c}{DAVIS 2017 val} \\
%\cline{5-10}
 & $\mathcal{J}$ \& $\mathcal{F}$ $\uparrow$ & $\mathcal{J}_M$ $\uparrow$ & $\mathcal{J}$ \& $\mathcal{F}$ $\uparrow$ & $\mathcal{J}_M$ $\uparrow$ \\
\midrule
   0.01   & 88.5 & 87.6 & 79.6 & 77.1 \\
   0.02  & 89.4 & 88.5 & 80.5 & 78.1 \\
   0.05  & \textbf{89.5} & \textbf{88.6} & \textbf{81.9} & \textbf{79.3} \\
   0.1  & 89.5 & 88.5 & 81.7 & 79.0 \\
   0.2 & 87.5 & 87.2 & 72.8 & 69.9 \\
   1 & 82.1 & 82.6 & 66.7 & 63.9 \\
\bottomrule
\end{tabular}
\vspace{-0.0 cm}
\end{center}
\caption{Ablation study on the temperature hyperparameter $\tau$.}
\label{tab:ablation_t}
\vspace{-0.3 cm}
\end{table}

%%%%%%%%%%%%%%%%%%%%%%%%%%%%%%%%%%%%%%%%%%%%%%%%%%%%%%%%%%%%%%%%%%%% Table Ablation of number of permutation-invariant features

\noindent{\textbf{The temperature hyperparameter $\tau$.}}  The similarity in this work is calculated by:
\begin{equation}
 \mathcal{K}(\mathbf{a}, \mathbf{b})=\exp(\frac{\mathbf{a}\mathbf{b}^{\top}/\tau}{\|\mathbf{a}\|\cdot \|\mathbf{b}\|}),
\label{eq:matching_tau}
\end{equation}
where the temperature hyperparameter $\tau$ controls the range of similarity measures. Table~\ref{tab:ablation_t} reports the segmentation performance on the DAIVS 2016 and DAVIS 2017 validation sets with different $\tau$. The segmentation performance is sensitive to $\tau$. Too large or too small $\tau$ are both not appropriate, especially the large one. The overall $\mathcal{J}\&\mathcal{F}$ on the DAVIS 2017 validation set is dropped from 81.9 to 66.7 when changing $\tau$ from 0.05 to 1. As contrast, $\tau=0.1$ and $\tau=0.05$ 0.05 is able to achieve satisfying results.

%%%%%%%%%%%%%%%%%%%%%%%%%%%%%%%%%%%%%%%%%%%%%%%%%%%%%%%%%%%%%%%%%%%% Table Ablation of tau

\begin{figure}[t]
\begin{center}
\includegraphics[width=1.\linewidth]{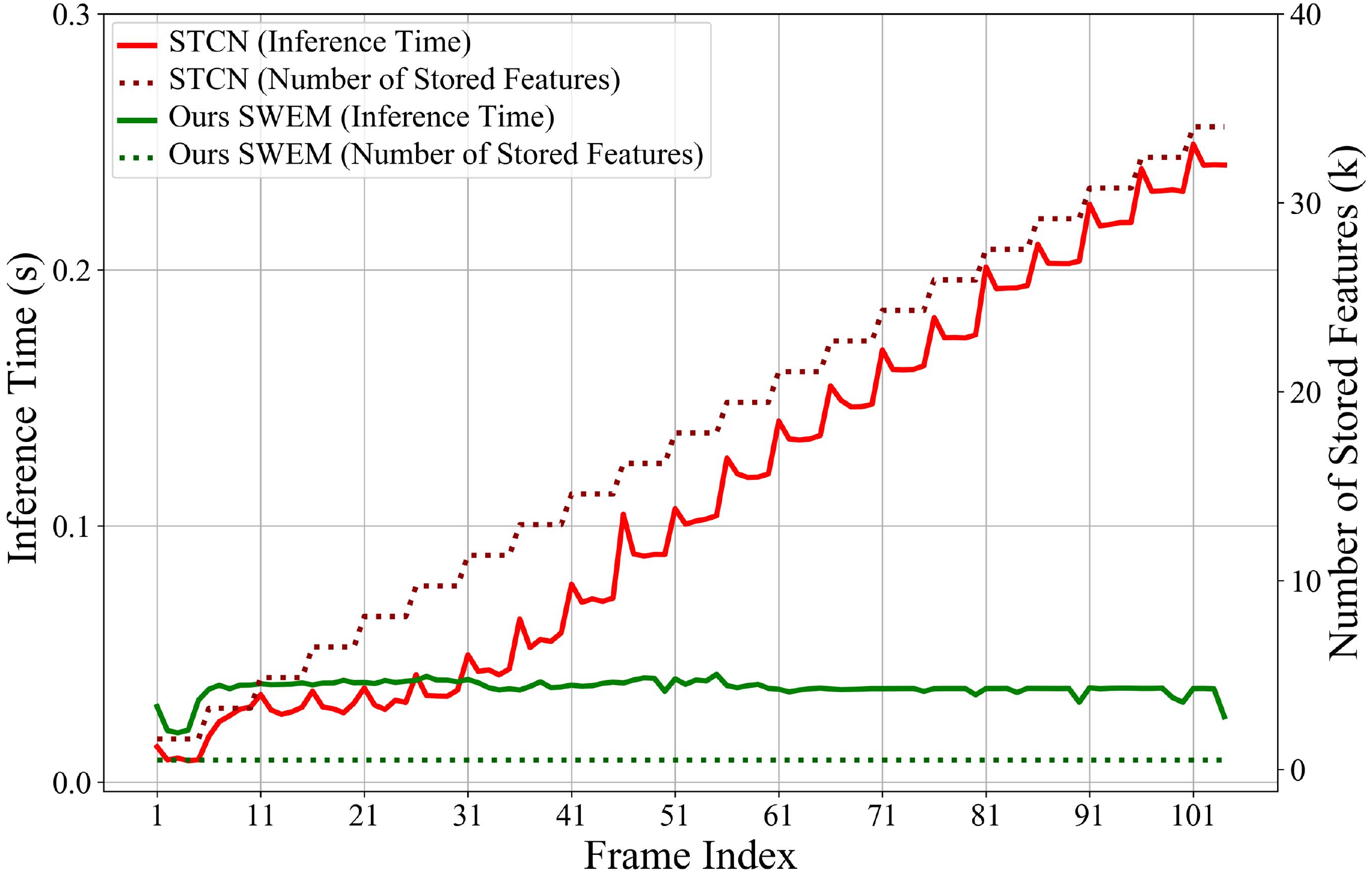}
\end{center}
%\vspace{-0.3 cm}
   \caption{The comparison between STCN and proposed SWEM on per-frame inference time and number of stored features in a single-object segmentation scenario. SWEM has a fixed size of memory features and stable inference time. Methods are evaluated with an NVIDIA GTX 1080ti GPU.}
\label{fig:Efficiency_davis16}
%\vspace{-0.2cm}
\end{figure}

\begin{figure}[b]
\begin{center}
\includegraphics[width=1.\linewidth]{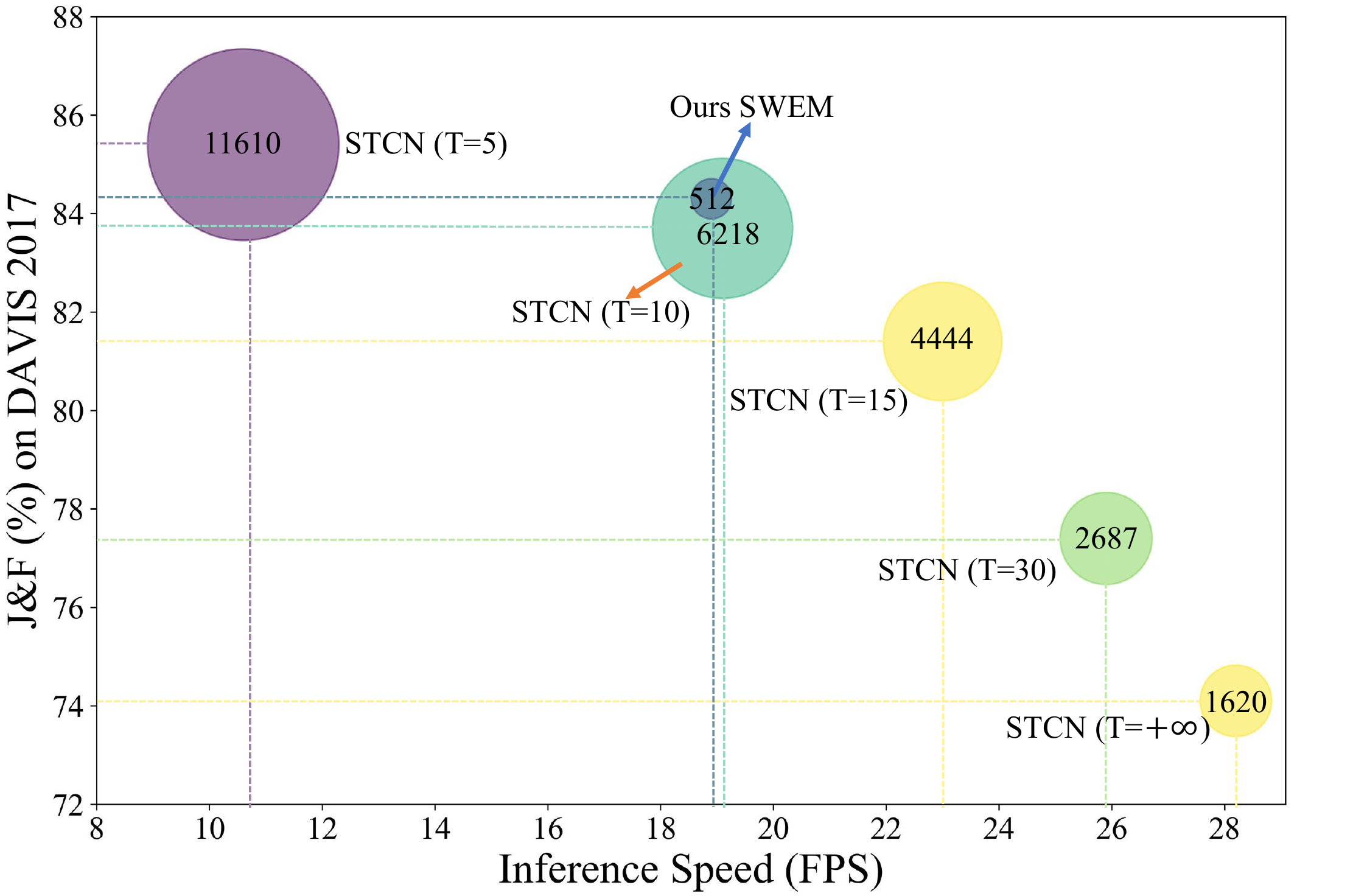}
\end{center}
%\vspace{-0.3 cm}
   \caption{The comparison between STCN and proposed SWEM on inference speed and performance in a multi-object segmentation scenario. We set different memory interval (T=5, 10, 15, 30, $+\infty$) for STCN. The area of the circle represents the number of per-frame memory features involved in matching for each object. SWEM achieves a better trade-off between performance and efficiency than STCN.}
\label{fig:Efficiency_davis17}
%\vspace{-0.2cm}
\end{figure}

\section{Efficiency comparison with STCN}
\label{sec:eff}
STCN~\cite{cheng2021stcn} achieves state-of-the-art performance while maintaining an acceptable inference speed (26 FPS with an NVIDIA V100 GPU). However, Similar to STM~\cite{Oh_2019_ICCV}, STCN still stores template features every T frames endlessly.
We compare the efficiency between STCN and proposed SWEM with an NVIDIA GTX 1080ti GPU, which is a cheap and commonly used platform. 

Figure~\ref{fig:Efficiency_davis16} shows the per-frame inference time and memory capacity of STCN and SWEM in a single-object segmentation scenario. It clearly reflects the negative effects of growing memory features on the efficiency and storage during the long-term segmentation. Compared to STCN, our SWEM stores far fewer memory features and has a stable inference time cost.
%%%%%%%%%%%%%%%%%%%%%%%%%%%%%%%%%%%%%%%%%%%%%%%%%%%%%%%%%%%%%%%%%%%% efficiency_swem_vs_stcn_davis16

Figure~\ref{fig:Efficiency_davis17} shows the efficiency and performance comparison between STCN and SWEM in a multi-object segmentation scenario. The area of the circle represents the number of per-frame per-object memory features. We also set different memory interval (T=5, 10, 15, 30, $+\infty$) for STCN. When $T=10$, STCN gets a similar inference speed and performance with SWEM, but with much more memory features. When $T=15$, the performance has a significant drop. If STCN only stores features of the first frame ($T=+\infty$), the overall performance is dropped from 85.4 to 74.1. 

The above results demonstrate the superiority of SWEM in terms of efficiency and performance with common hardware.

%%%%%%%%%%%%%%%%%%%%%%%%%%%%%%%%%%%%%%%%%%%%%%%%%%%%%%%%%%%%%%%%%%%% Table SOTAs on YTVOS19
\begin{table}[b]
\begin{center}
\begin{tabular}{cccccc}
\toprule 
  \multirow{2}*{Method} &\multirow{2}*{$\mathcal{G}$} & \multicolumn{2}{c}{seen} & \multicolumn{2}{c}{unseen} \\
 & & $\mathcal{J}_M$ $\uparrow$& $\mathcal{F}_M$ $\uparrow$ & $\mathcal{J}_M$ $\uparrow$ & $\mathcal{F}_M$ $\uparrow$\\
\midrule
STM~\cite{Oh_2019_ICCV}                & 79.2 & 79.6 & 83.6 & 73.0 & 80.6 \\
CFBI~\cite{yang_2020_ECCV}             & 81.0 & 80.6 & 85.1 & 75.2 & 83.0 \\
LWL~\cite{Bhat_2020_ECCV_GCM}          & 81.0 & 79.6 & 83.8 & 76.4 & 84.2 \\
SSTVOS~\cite{duke2021sstvos}           & 81.8 & 80.9 & 85.3 & 76.6 & 84.4 \\
HMMN~\cite{seong2021hierarchical}      & 82.5 & 81.7 & 86.1 & 77.3 & 85.0 \\
JOINT~\cite{mao2021joint}              & 82.8 & 80.8 & 84.8 & \textbf{79.0} & 86.6 \\
AOT~\cite{yang2021associating}         & \textbf{83.6} & \textbf{82.2} & \textbf{86.9} & 78.3 & \textbf{86.9} \\
\midrule
\textbf{SWEM*}                         & 82.6 & 82.0 & 86.1 & 77.2 & 85.2 \\
\bottomrule
\end{tabular}
\end{center}
\vspace{-0.0 cm}
\caption{Comparison with state-of-the-art methods on the YouTube-VOS 2019 validation dataset. We report all of the mean Jaccard ($\mathcal{J}$), the boundary ($\mathcal{F}$) scores for seen and unseen categories, and the overall scores $\mathcal{G}$. Besides, we use `*' to indicate those methods with an inference speed $>20$ FPS. Note SSTVOS, JOINT and AOT are transformer-based methods.}
\label{tab:YTVOS19}
\vspace{-0.0 cm}
\end{table}
%%%%%%%%%%%%%%%%%%%%%%%%%%%%%%%%%%%%%%%%%%%%%%%%%%%%%%%%%%%%%%%%%%%% Table SOTAs on YTVOS19

%%%%%%%%%%%%%%%%%%%%%%%%%%%%%%%%%%%%%%%%%%%%%%%%%%%%%%%%%%%%%%%%%%%% 
\begin{table*}[t]
\begin{center}
\begin{tabular}{ccccccc}
\toprule
 \multirow{2}*{\diagbox[width=10em]{Training Data}{Method}} & \multicolumn{3}{c}{DAVIS 2016 val $\mathcal{J}$ \& $\mathcal{F}$} & \multicolumn{3}{c}{DAVIS 2017 val $\mathcal{J}$ \& $\mathcal{F}$} \\
%\cline{5-10}
 & STM~\cite{Oh_2019_ICCV} & STCN~\cite{cheng2021stcn} & \textbf{SWEM} & STM~\cite{Oh_2019_ICCV} & STCN~\cite{cheng2021stcn} & \textbf{SWEM}\\
\midrule
DAVIS                       & 74.5 & 84.6 & \textbf{88.1}      & 48.6 & 71.2 & \textbf{77.2} \\
DAVIS + YTVOS               & 88.2 & \textbf{89.8} & 89.5      & 80.0 & 81.1 & \textbf{81.9} \\
Image + (DAVIS + YTVOS)     & 89.8 & 91.2 & \textbf{91.3}      & 81.6 & \textbf{84.5} & 84.3 \\

\bottomrule
\end{tabular}
\end{center}
\caption{Analysis of training on different datasets. We train STM~\cite{Oh_2019_ICCV}, STCN~\cite{cheng2021stcn} and proposed SWEM under three different conditions: 1) only training on video data DAVIS, 2) training on video datasets DAVIS and YouTube-VOS, and 3) pretraining on image data first and then training on DAVIS and YouTube-VOS.}
\label{tab:ablation_data}
\end{table*}

\section{Comparisons on YouTube-VOS 2019}
\label{sec:YTVOS19}
To further evaluate the effectiveness of our method, we
also carry out experiments on the YouTube-VOS 2019 dataset.
Note that the YouTube-VOS 2019 validation set contains 507 videos, while the 2018 version has 474 videos.
Specifically, our SWEM is trained on the YouTube-VOS 2019 training set and evaluated on the validation set through the official evaluation server. We compare SWEM with recent state-of-the-art methods in Table~\ref{tab:YTVOS19}, including transformer-based methods SSTVOS~\cite{duke2021sstvos}, JOINT~\cite{mao2021joint} and AOT~\cite{yang2021associating}. SWEM achieves the 82.6\% overall score while maintaining the real-time inference speed.

\section{Impact of Training Data}
\label{sec:TrainingData}
Table ~\ref{tab:ablation_data} presents the performance of various models on DAVIS 2016 and DAVIS 2017 validation sets with different training datasets. 
It can be seen that both pre-training on image data and using additional YouTube-VOS video data can boost performance. Note that our SWEM trained only on DAVIS data outperforms STM~\cite{Oh_2019_ICCV} and STCN~\cite{cheng2021stcn} under the same setting with a large margin. Furthermore, STM~\cite{Oh_2019_ICCV} obtains $48.6\%$ $\mathcal{J}\&\mathcal{F}$ on DAVIS 2017 validation set when it is trained only with DAVIS data. Authors claim it is due to the high risk of over-fitting on small datasets. However, STCN gets a much higher overall performance $71.2\%$ $\mathcal{J}\&\mathcal{F}$ than STM. We argue the reason for this performance gap is that the features involved in matching in STM come from different feature spaces, whereas those features in STCN and SWEM are from the same feature spaces. Moreover, the direct usage of affinity features also accelerates the convergence of SWEM. SWEM gets similar performance with STCN when adopting additional YouTube-VOS video data or image data. It is worth noting that SWEM stores fixed-size (512) memory features which are much fewer than those in STCN. As an inference, a single frame with a size $480\times 864$ produces 1,620 features.

\section{Qualitative Results}
\label{sec:qualitative}
\begin{figure*}[h]
\begin{center}
\includegraphics[width=0.95\linewidth]{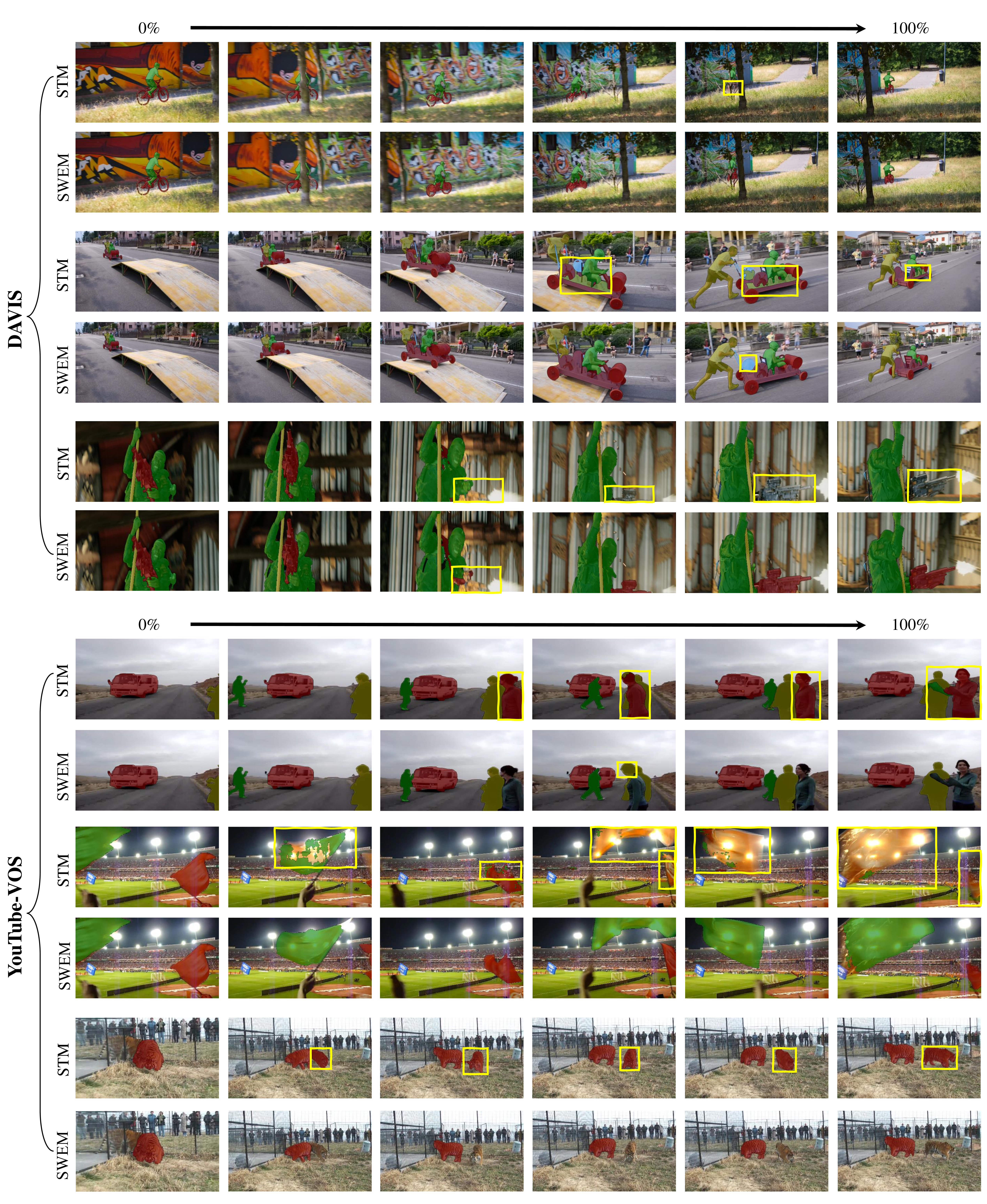}
\end{center}
\vspace{-0.0 cm}
   \caption{The qualitative comparison between STM~\cite{Oh_2019_ICCV} and our SWEM on the DAVIS 2017 and YouTube-VOS. The obvious failure segmentation is indicated by yellow bounding boxes. Our SWEM is robust with rapid motion and similar distractors.}
\label{fig:vis}
\vspace{-0.0 cm}
\end{figure*}

We further provide the qualitative comparison between STM~\cite{Oh_2019_ICCV} and our SWEM on the validation set of the DAVIS 2017 and YouTube-VOS in Figure~\ref{fig:vis}, which demonstrates the superiority of our SWEM. 

%\end{appendices}

\end{document}